\documentclass{article}

\usepackage{microtype}
\usepackage{graphicx}
\usepackage{subfigure}
\usepackage{booktabs} % for professional tables

\usepackage{algorithm}
\usepackage{algorithmic}
\usepackage{amsbsy}
\usepackage{verbatim}
\usepackage{multirow}
\usepackage{amssymb}
\usepackage{amsmath}
\usepackage{pifont}
\usepackage{xcolor}
\usepackage{caption}
\usepackage{adjustbox}
\usepackage{hyperref}

\newcommand{\ie}{\emph{i.e.}}    % notation of `i.e.`
\newcommand{\eg}{\emph{e.g.}}    % notation of `e.g.`
\usepackage[accepted]{icml2023_arxiv}

\usepackage{mathtools}
\usepackage{amsthm}

\usepackage[capitalize,noabbrev]{cleveref}

\theoremstyle{plain}

\theoremstyle{definition}

\theoremstyle{remark}

\usepackage[textsize=tiny]{todonotes}

\icmltitlerunning{Crafting Training Degradation Distribution for the Accuracy-Generalization Trade-off}

\begin{document}

\twocolumn[
\icmltitle{Crafting Training Degradation Distribution for the Accuracy-Generalization Trade-off in Real-World Super-Resolution}



\icmlsetsymbol{equal}{*}

\begin{icmlauthorlist}
\icmlauthor{Ruofan Zhang}{thu}
\icmlauthor{Jinjin Gu}{shlab,usyd}
\icmlauthor{Haoyu Chen}{hkustgz}
\icmlauthor{Chao Dong}{shlab,siat}
\icmlauthor{Yulun Zhang}{eth}
\icmlauthor{Wenming Yang}{thu}
\end{icmlauthorlist}

\icmlaffiliation{thu}{Tsinghua Shenzhen International Graduate School, Tsinghua University}
\icmlaffiliation{usyd}{The University of Sydney}
\icmlaffiliation{shlab}{Shanghai AI Laboratory}
\icmlaffiliation{siat}{Shenzhen Institutes of Advanced Technology, Chinese Academy of Sciences}
\icmlaffiliation{hkustgz}{The Hong Kong University of Science and Technology (Guangzhou)}
\icmlaffiliation{eth}{ETH Z\"{u}rich}
\icmlcorrespondingauthor{Jinjin Gu}{jinjin.gu@sydney.edu.au}
\icmlcorrespondingauthor{Wenming Yang}{yang.wenming@sz.tsinghua.edu.cn}

\icmlkeywords{Machine Learning, ICML}

\vskip 0.3in
]



\printAffiliationsAndNotice{} % otherwise use the standard text.

\begin{abstract}
Super-resolution (SR) techniques designed for real-world applications commonly encounter two primary challenges: generalization performance and restoration accuracy.
We demonstrate that when methods are trained using complex, large-range degradations to enhance generalization, a decline in accuracy is inevitable.
However, since the degradation in a certain real-world applications typically exhibits a limited variation range, it becomes feasible to strike a trade-off between generalization performance and testing accuracy within this scope.
In this work, we introduce a novel approach to craft training degradation distributions using a small set of reference images.
Our strategy is founded upon the binned representation of the degradation space and the Fréchet distance between degradation distributions.
Our results indicate that the proposed technique significantly improves the performance of test images while preserving generalization capabilities in real-world applications.
\end{abstract}

\vspace{-4mm}
\section{Introduction}

\begin{figure}[t]
    \centering
   \includegraphics[width=\linewidth]{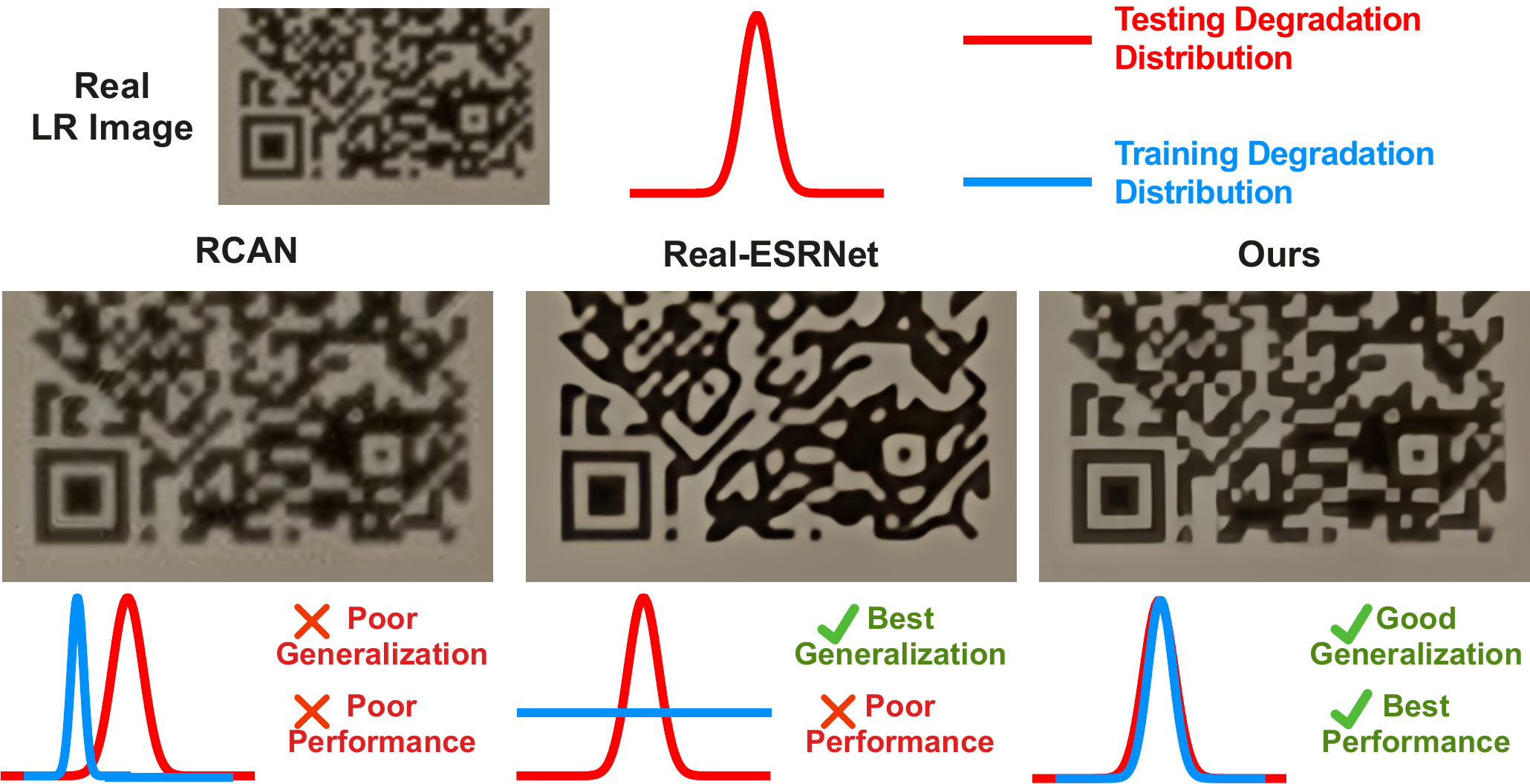}
   \vspace{-5mm}
   \caption{The figure shows the different effects of different training degradation distributions (shown in blue lines) on the target test distribution (shown in red lines). (a) The generalization performance of SR models limits their application when the training distributions are insufficient or mismatched. (b) When the training distribution is too large, the generalization of the SR model is better, but the overall accuracy will drop dramatically. (c) The proposed method improves the accuracy of test images as much as possible while ensuring the model's generalization performance.}
    \label{fig:teaser}
    \vspace{-6mm}
\end{figure}

Image Super-Resolution (SR) is focused on reconstructing high-resolution (HR) images from their corresponding low-resolution (LR) or degraded observations.
SR has a rich history of utilizing deep learning techniques, dating back to the groundbreaking work of SRCNN~\cite{srcnn2014}.
With the advanced modeling capacity of deep networks, the performance of SR networks has experienced rapid progress.
Nevertheless, it is widely recognized that the efficacy of SR models in practical applications is heavily influenced by their generalization performance and the training degradations~\cite{liu2022blind}.
The complex and diverse degradation scenarios encountered in real-world applications present considerable challenges to the successful implementation of SR techniques.

A potential solution to tackle real-world SR challenges is the adoption of blind SR methods.
These approaches generally rely on a predefined degradation model, while assuming that certain parameters remain unknown, such as the extent of downsampling blur or noise level.
Consequently, blind SR methods are capable of addressing SR problems within a specific degradation range.
However, the utilization of predefined degradation models limits their applicability to a narrow range of degradations, rendering them incapable of generalizing to mixed, intricate degradation cases encountered in real-world applications.
As a result, these methods continue to exhibit sub-optimal generalization performance.

Recently, a new SR paradigm that employs a vast array of complex or mixed degradation data for training has increasingly gained interest within the research community.
Notable methods include BSRGAN \cite{zhang2021designing} and Real-ESRGAN \cite{wang2021real}.
By leveraging large and diverse synthetic training datasets in an end-to-end manner, these techniques achieve substantial generalization performance.
In practice, the degradation distribution used for training is determined by a manually configured stochastic process.
Generally, the distribution range is set to be extensive in order to enhance generalization performance across degradations.
However, there are inevitable trade-offs.
In exchange for improved generalization performance, the accuracy of these models experiences a significant decline.
Common issues include the removal of texture details, the generation of blurred or inaccurate edges, and the unwarranted sharpening of blurred backgrounds.

In this paper, we investigate the trade-off between accuracy and generalization from a more pragmatic standpoint.
We contend that in the majority of application scenarios, degradation range tends to be limited, for instance, images captured by a specific type of image sensor, different frames from the same old movie, or old photos originating from the same era.
Although these degradations are complex, which renders blind methods based on predefined degradation models ineffective, the range of degradation variation is relatively small compared to existing data synthesis strategies.
The excessive portion of training degradation in conventional practices adversely impacts SR performance within the target range.
Given the availability of images requiring super-resolution in practical applications (despite the unknown degradation process responsible for these images), we can customize the degradation distribution used for training to better suit the target test images.
We illustrate this process in \figurename~\ref{fig:teaser}.
In summary, we explore a novel SR problem wherein, \emph{given access to some test images, the training degradation distribution is modified to enhance performance within the target test degradation range while preserving generalization performance}.

To address this novel problem, we introduce two primary technical designs.
Firstly, to determine a distribution within the space of potential degradations and sample training degradations from it, we require a suitable representation of the degradation distribution.
We employ the binning method to partition the feasible degradation space into multiple distinct intervals.
Uniform sampling is conducted within an interval, while weighted sampling is performed between different intervals (bins).
This approach allows us to obtain a simple, parameterized sampling method for the degradation space.
Secondly, we propose measuring the distance between two degradation distributions by calculating the Fréchet distance of deep features, even when the content of the two image sets differs.
Based on the obtained degradation distribution distance, we calculate the weight for the degradation distribution bins, thereby parameterizing the training degradation distribution.
We conduct extensive experiments for the proposed SR problem and method using both synthetic data and real-world low-resolution images.
Our method demonstrates robust quantitative performance while maintaining strong generalization capabilities.

\vspace{-2mm}
\section{Related Work}
\vspace{-2mm}
\paragraph{Super-Resolution.}
Single image super-resolution (SR), which aims at reconstructing a high-resolution (HR) image from its low-resolution (LR) observations, is a long-standing problem in the low-level vision field.
Since SRCNN \cite{srcnn2014}, which is the pioneering work of employing deep learning in SR, deep SR networks have brought prosperous development in this field.
Plenty of deep learning based SR methods have been proposed, including deeper networks \cite{vdsr2016,espcn2016}, light-weight networks \cite{fsrcnn2016,gu2022super,zhou2023efficient}, recurrent architectures \cite{drcn2016,drrn2017}, residual architectures \cite{srgan2017,wang2018esrgan,li2022blueprint}, attention networks \cite{rcan2018,san2019}, and Transformer networks \cite{chen2021pre,chen2022cross,liang2021swinir,chen2023activating}.
However, most of these methods are aimed at the laboratory environment with pre-defined degradations, and the effect is limited in real applications.

\vspace{-3mm}
\paragraph{Blind SR} methods
are proposed to solve the problem of SR model failure in real-world applications.
The community has already reached a relatively clear conclusion for the reasons of the failure, that is, the mismatch between training and testing degradations \cite{liu2022blind}.
Early blind SR methods usually assume a pre-defined degradation model with some unknown parameters, \eg, the parameters of the blur kernel and the noise level \cite{Gu_2019_CVPR,huang2020unfolding,cornillere2019blind}.
These methods still fail in a wide range of complex situations because real-world degradations are very complex.
Simple degradation models cannot represent these situations.
And the generalization ability of these methods is also not enough to make them applicable in the wild.
Then, methods with implicit modeling are proposed and do not depend on any explicit parameterization.
Utilizing data distribution within the external datasets, they often learn the underlying SR model implicitly, \eg, CinCGAN \cite{yuan2018unsupervised}, DASR \cite{wang2021unsupervised}, FS-SRGAN \cite{zhou2020guided}, and FSSR \cite{fritsche2019frequency}.
More recently, these methods have evolved further, with successful training on complex, large-range degradation data.
\citet{wang2021real} propose a novel data synthesizing method called high-order degradation model and train Real-ESRGAN.
\citet{zhang2021designing} propose BSRGAN that randomly applies different degradation operations during data synthesizing.
Despite progress in visual effects, these methods rely on handset training degradation distributions.
Both Real-ESRGAN and BSRGAN suffer from significant accuracy degradation when the training distribution is too wide.

More related to this work, there are also methods that consider the case where a part of the test input image can be obtained as a reference.
\citet{wang2021domain} propose an unpaired SR training framework based on feature domain adaptation.
\citet{luo2022learning} use adversarial training methods to generate specific training degradations.

\vspace{-3mm}
\paragraph{Generalization Performance of SR models.}
The generalization performance of an SR network is crucial for its effectiveness on unseen data \cite{chen2023masked,gu2023networks}.
To date, limited research has focused on explaining, evaluating, and improving the generalization performance of SR networks.
One study \citet{liu2021discovering} discovered that SR networks tend to overfit to degradations and exhibit characteristic degradation ``semantics'' (DDR) within the network, which often leads to a decrease in generalization ability.
Building on this finding, another study developed a generalization assessment index for SR networks called SRGA \citet{liu2022evaluating}.
This non-parametric, non-learning metric measures generalization ability by examining the statistical characteristics of deep features within SR networks, rather than output images.
As the generalization performance of SR gains increasing attention, specialized methods for enhancing SR generalization performance have emerged.
For instance, one study demonstrated that the appropriate use of dropout benefits SR networks and improves generalization ability \citet{kong2022reflash}.
The goal of this paper is to explore ways to achieve higher accuracy while ensuring robust generalization performance.

\vspace{-2mm}
\section{Methodology}
\vspace{-1mm}
\subsection{Problem Formulation}
\vspace{-2mm}
The proposed SR problem can be formulated as follows.
Suppose we are designing an SR model for a new application, and we have obtained a set of reference degraded images $X_{ref}=\{x^r_i\}_{i=1}^n$ relevant to the application, and a set of test degraded images $X_{test}=\{x^t_i\}_{i=1}^N$ for evaluation.
We can assume that these images are generated from the corresponding high-quality images $Y_{ref}$ and $Y_{test}$ with different degradations sampled from the same degradation distribution $P_r(d)$, where $d$ denotes a random degradation and $P_r$ represents the testing degradation distribution.
This process can be formulated as $X=\mathcal{D}(Y, P_r)$, indicating that $x_i = d(y_i)$ for $x_i\in X$, $y_i\in Y$, and $d\sim P_r(d)$.

We now introduce a new degradation distribution $P_t$ for training. 
Given a set of high-quality training images $Y_{trn}$, we synthesize the training degraded images $X_{trn}=\mathcal{D}(Y_{trn}, P_t)$ and then obtained the SR model $F_\theta$.
Here, $\theta$ is determined by $P_t$ via a conditional distribution $\theta\sim P(\theta|P_t)$, as different training data will produce different SR models.
Our goal is to maximize the performance of the obtained SR model $F_\theta$ on the target test degradation $P_r$ by changing $P_t$. This can be formulated as:
\begin{equation}
    \max_{P_t} \mathbb{E}_{\theta\sim P(\theta|P_t)} \mathcal{S}(F_{\theta}(\mathcal{D}(Y,P_r)),Y),
\end{equation}
where $\mathcal{S}$ represents an image similarity metric used to evaluate the image reconstruction, \eg, PSNR, SSIM.

\vspace{-2mm}
\subsection{Motivation}
\begin{figure}[t]
    \centering
   \includegraphics[width=\linewidth]{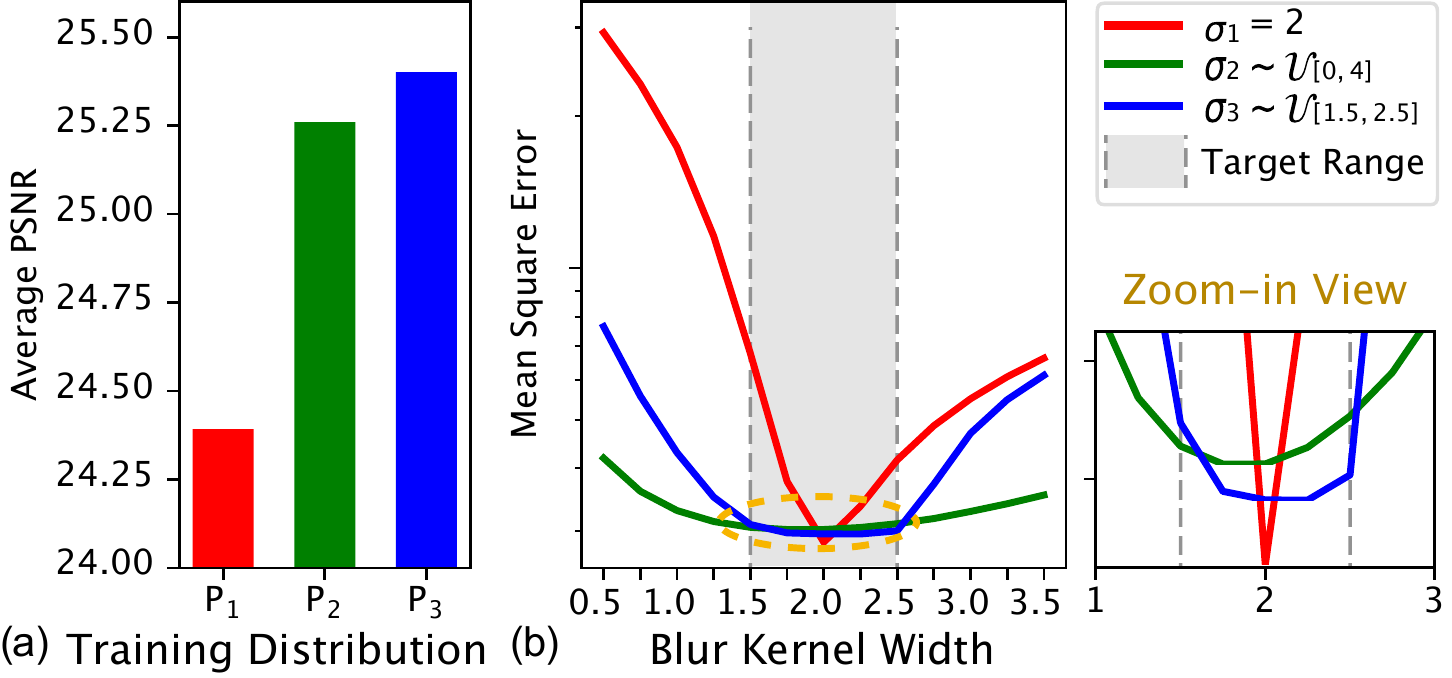}
   \vspace{-7mm}
   \caption{(a) shows the average PSNR performance. (b) shows the performance under different blur conditions. These figures show the importance of the training with appropriate degradation distribution $P_t$.}
    \label{fig:motivation}
    \vspace{-4mm}
\end{figure}

\vspace{-2mm}
We then review the importance of the training with appropriate degradation distribution $P_t$ using experiments on synthetic data.
We assume a simple degradation model $x=d(y)=(y\otimes k)\downarrow$, where $k$ is the blur kernel and $\downarrow$ denotes downsampling.
We set $P_r$ as the degradation distribution formed by sampling blur kernel widths $\sigma_r$ from 1.5 to 2.5 uniformly, denoted as $\sigma_r\sim\mathcal{U}_{[1.5,2.5]}$.
We train three SR models with different training distributions $P_1$, $P_2$ and $P_3$ and observe their behavior, where $\sigma_1=2$, $\sigma_2\sim\mathcal{U}_{[0,4]}$ and $\sigma_3\sim\mathcal{U}_{[1.5,2.5]}$.
The testing average PSNR is shown in \figurename~\ref{fig:motivation} (a).
We also test the performance of these models under different blur conditions separately, shown in \figurename~\ref{fig:motivation} (b).
It can be seen that the model trained with a single degradation cannot handle other degradations within $P_r$ except the training degradation.
The SR model trained with $P_2$ degradation distribution can handle a large range of blurs, and is similar to the existing practices such as Real-ESRGAN and BSRGAN.
This approach does bring about excellent generalization ability even beyond the target blur range, but in order to take into account a larger range of inputs, the restoration accuracy of all degradations in the range is reduced when the network capacity remains unchanged.
The final model uses a matched training distribution and achieves the best PSNR performance while maintaining its generalization performance in the target range.

This experiment justifies our problem from three aspects:
First, training within a certain degradation \emph{range} is necessary because it can provide generalization performance to avoid performance drop when the degradation model is slightly mismatched.
Second, this range is not the bigger, the better. An SR model training with a larger range may generalize to more images but suffer a performance drop on all images.
And third, the closer the training degradation distribution $P_t$ is to the test degradation distribution $P_r$, the better the final result.
We next describe the technical designs to achieve our goal.

\vspace{-2mm}
\subsection{Degradation Distribution Binning and Sampling}
\vspace{-2mm}
The representation and quantification of the degradation distribution are essential to determine the training distribution $P_t$.
Existing methods represent the degradation process as a pipeline, \eg, the high-order degradation in Real-ESRGAN.
In the pipeline, operations such as random blurring, noise, and compression are sequentially performed on the image.
The parameter for each operation is given by a pre-set distribution.
This practice makes it extremely difficult to quantify and control its distribution.
First, its different operations are performed in sequence, and the previous operations will affect the subsequent operations and get very different results.
If we want to update the parameter distribution, it is difficult for us to attribute the contribution of each step in the degradation.
Second, its degradation parameters follow the continuous distributions, which poses challenges for us in conducting statistical estimation or inference.

We propose a binning methodology for discretizing the joint distribution of multiple degradation parameters. This simplifies the representation and sampling process of the degradation distribution.
We describe our binning method based on a widely used image degradation model:
\begin{equation}
    x =\mathcal{D}(y, d)= \mathcal{C}_q\circ\mathcal{E}_l\circ\mathcal{B}_\sigma(y)=[(y\otimes k)\downarrow+\epsilon]_{\mathtt{JPEG}},
\end{equation}
where $\sigma$, $l$, and $q$ are parameters, \ie, kernel width for blurring $\mathcal{B}$, noise level for noising $\mathcal{E}$ and quality level for compression $\mathcal{C}$.
The distribution of degradation is described by a joint distribution of these parameters $P(d)=p(q, l, \sigma)$.
Typically, these parameters are sampled from uniform distributions.
In our work, $\sigma\sim\mathcal{U}_{[0,5]}$, $l\sim\mathcal{U}_{[0,50]}$ and $q\sim\mathcal{U}_{[30,90]}$.
We divide these continuous distributions evenly into discrete intervals.
For example, the compression parameter $q$ is partitioned into three bins as $q_1\sim\mathcal{U}_{[30,50]}$, $q_2\sim\mathcal{U}_{[50,70]}$ and $q_3\sim\mathcal{U}_{[70,90]}$.
We use the same binning method for noise and blur and divide it into five equal bins, respectively.
Binning these three components of degradation yields $3\times5\times5=75$ possible degradation bins.
Then sampling from the entire degradation space can be viewed as first sampling a bin from the set of bins, and then sampling a degradation from this bin.
It can be formulated as $P(d)=P(q, l, \sigma)=\sum_bp(q, l, \sigma|b)p(b)$, where $b$ is the random variable of the bin and $p(b)$ is its distribution.

\begin{figure}[t]
    \centering
   \includegraphics[width=\linewidth]{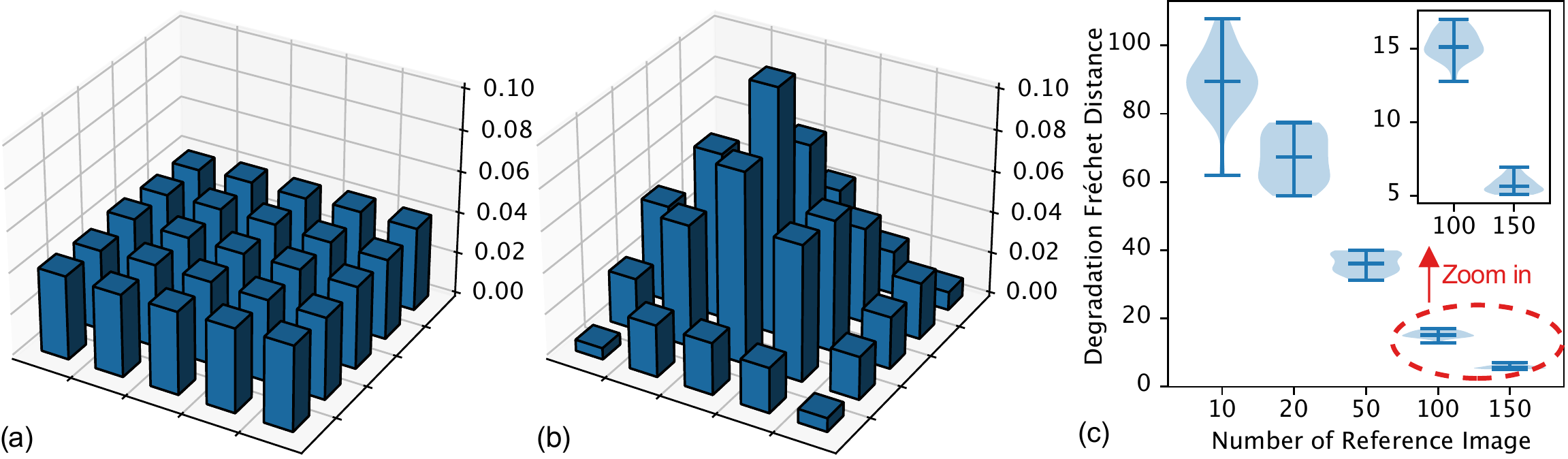}
   \vspace{-7mm}
   \caption{(a) Perform binning operations on multiple degradation parameters and assign uniform sampling weights at the beginning. (b) Updating the sampling weights of the bins in order to sample in the joint distribution of multiple degradation parameters. (c) The estimated degradation distance using different amounts of images.}
    \label{fig:box_curve}
    \vspace{-4mm}
\end{figure}

We next change the distribution of the degradations $P(d)$ by changing $p(b)$.
Mathematically, we assign a sampling weight (importance) to each bin and formulate $p(b)$ as $p(b_i)=w_i$ for $i\in\{1, 2, \dots, N_{bin}\}$, where $N_{bin}$ is the number of bins.
In the initial stage, we give each bin the same uniform sampling weight, and the result of sampling at this time is equivalent to uniform sampling over the entire interval.
By updating the weight vector $\mathbf{w}\in\mathbb{R}^{N_{bin}}$, we can shape the degradation distribution $P(d)$.
\figurename~\ref{fig:box_curve} (a) and (b) show a schematic illustration of this process.
It is worth noting that when the number of steps in the degradation process increases, such as using the high-order degradation model, the number of binning increases exponentially.
However, we found that lower-order degradation models can already provide good generalization performance and cover most situations, as long as their distributions are well-matched.

\vspace{-2mm}
\subsection{Obtaining Weight Using Feature Fréchet Distance}
\vspace{-2mm}
We next describe the method to obtain this binning sampling weight given a set of real degraded images $X_{ref}$.
Recall that in the studied problem, we have access to a small number of real degraded images as references to help estimate the test degradation distribution $P_r$.
Since we only need to adjust $\mathbf{w}$, the probability of sampling from each bin, we need to calculate a kind of distance between the degradation $p(d|b)$ in each bin $b$ and the degradation of $P_r$.
The main difficulty is that we don't have high-quality images with the same content as the $X_{ref}$.
Thus, we introduce another set $Y_{w}=\{y^w_i\}_{i=1}^n$ of high-quality images to synthesize degraded images $X_w^b=\mathcal{D}(Y_{w}, p(d|b))$ according to the degradation in each bin $b$.
Then the distance of $p(d|b)$ to $P_r$ is given by the Fréchet distance \cite{frechet1957distance} between the deep features $\phi(X_w^b)$ and $\phi(X_{ref})$ of $X_w^b$ and $X_{ref}$, where $\phi(\cdot)$ is a deep feature extractor, and the features are of size $\mathbb{R}^{n\times c}$ with feature dimension $c$.
Following the common practice of using the Fréchet distance to measure the distance between two deep features \cite{heusel2017gans}, we assume that these $c$ dimensional deep features follow their respective $c$ dimensional Gaussian distributions.
We give the solution of the Fréchet distance between them as
\begin{align}
    &D_F(\phi(X_w^b), \phi(X_{ref}))^2=\\&\|\mu_b-\mu_{ref}\|^2_2
    +\mathbf{tr}\big(\Sigma_{b}+\Sigma_{ref}-2(\Sigma_b^{\frac{1}{2}}\Sigma_{ref}\Sigma_b^{\frac{1}{2}})^{\frac{1}{2}}\big),\notag
\end{align}
where $\mathcal{N}(\mu_b,\Sigma_b)$ is the fitted Gaussian distribution using $\phi(X_w^b)$ and $\mathcal{N}(\mu_{ref},\Sigma_{ref})$ is fitted using $\phi(X_{ref})$.

We calculate this distance for all the bins and obtain $\mathbf{D}\in\mathbb{R}^{N_{bin}}$, where $\mathbf{D}_i=D_F(\phi(X_w^{bi}), \phi(X_{ref}))^2$.
We first normalize the vector $\mathbf{D}_{norm}$ linearly into interval $[0, 1]$.
We assign sampling weights to these bins based on $\mathbf{D}_{norm}$ using the following function:
\begin{equation}
    w_i=\frac{\exp((1-\mathbf{D}_{norm}[i])^\alpha)-1}{\sum_{j=1}^{N_{bin}}[\exp((1-\mathbf{D}_{norm}[j])^\alpha)-1]},
    \label{eq:softmax}
\end{equation}
where $\alpha$ is a hyper-parameter that controls the kurtosis of the distribution.
When $\alpha$ is set to be a large value, the resulting distribution will be concentrated in a small range.
On the contrary, the formed distribution will be wider, and it degenerates to a uniform distribution when $\alpha=0$.
Equation \eqref{eq:softmax} also ensures that $\sum_i^{N_{bin}} p(b_i)=1$.

\begin{figure*}[t]
    \centering
    
    % \resizebox{0.35\textwidth}{!}{
    \raisebox{80pt}{
    \begin{minipage}[c]{0.18\textwidth}
        \includegraphics[width=\linewidth]{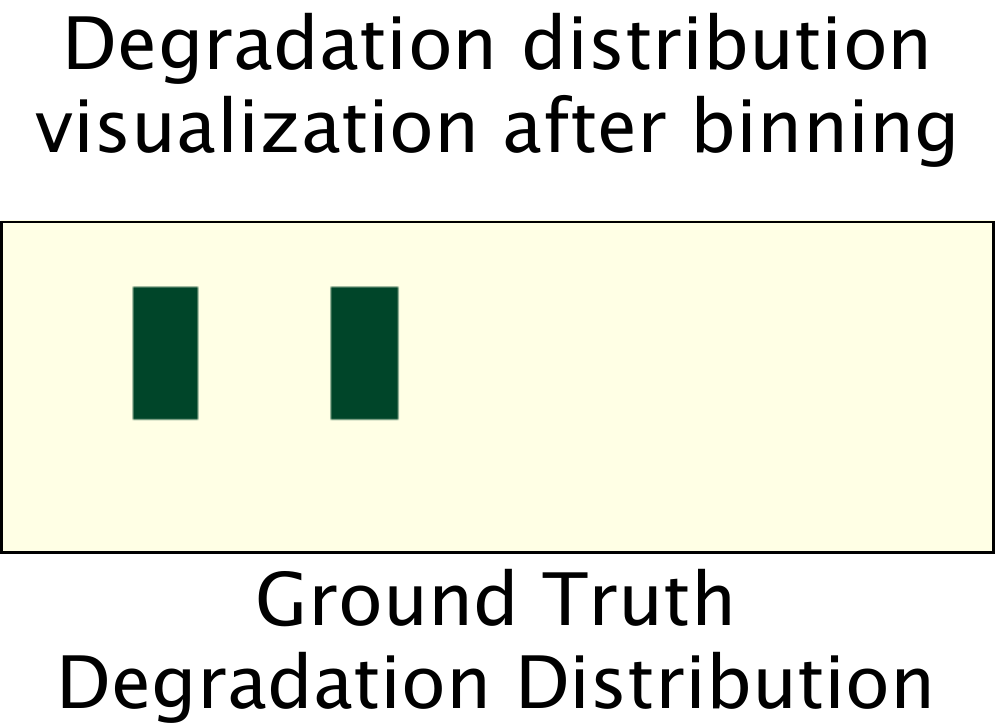}
        \caption{The visualization of the estimated sampling weight when using different $\alpha$s and feature extractors.}
	   \label{fig:alpha}
    \end{minipage}
    }
    \hfill 
    \resizebox{0.8\textwidth}{!}{
        \includegraphics[width=\linewidth]{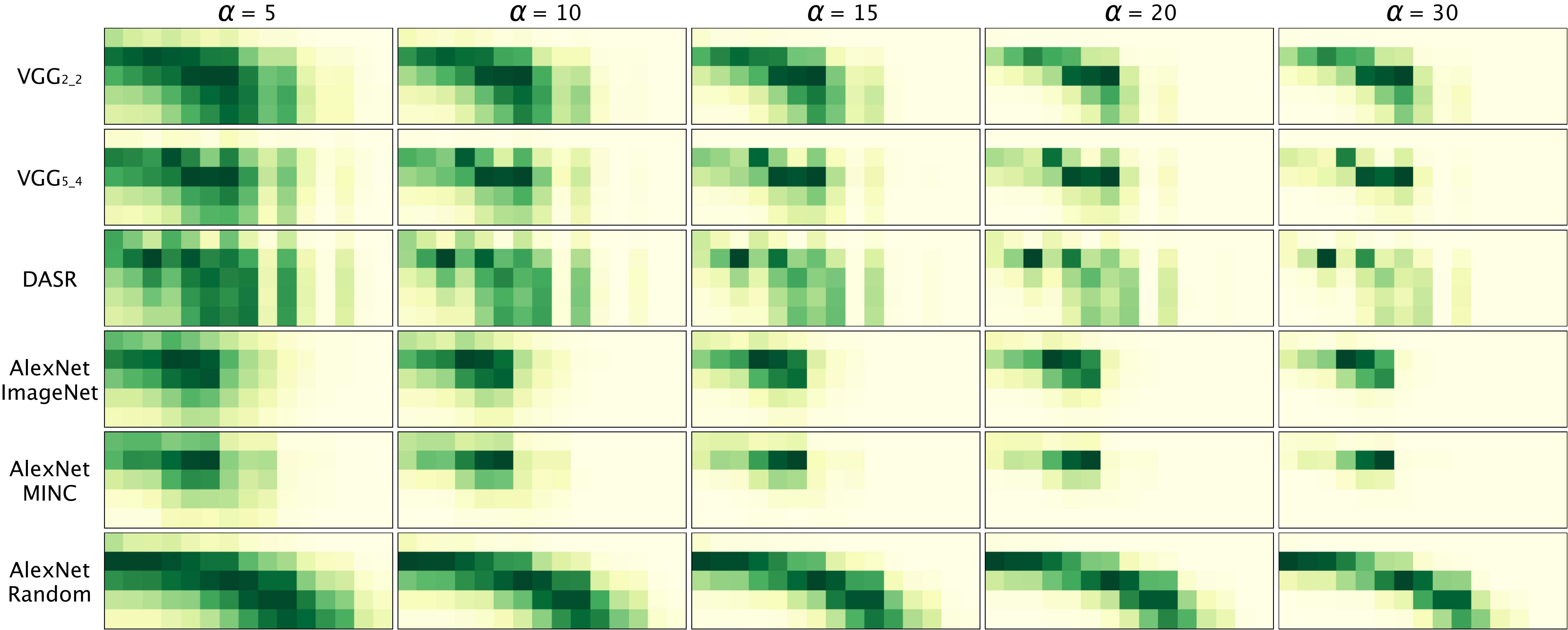}
    }
    % }
    \vspace{-2mm}
\end{figure*}

\vspace{-2mm}
\subsection{Discussion}
\vspace{-2mm}
We have gotten the full picture of our solution, but some issues are still worth discussing.

\begin{figure}[t]
    \centering
   \includegraphics[width=\linewidth]{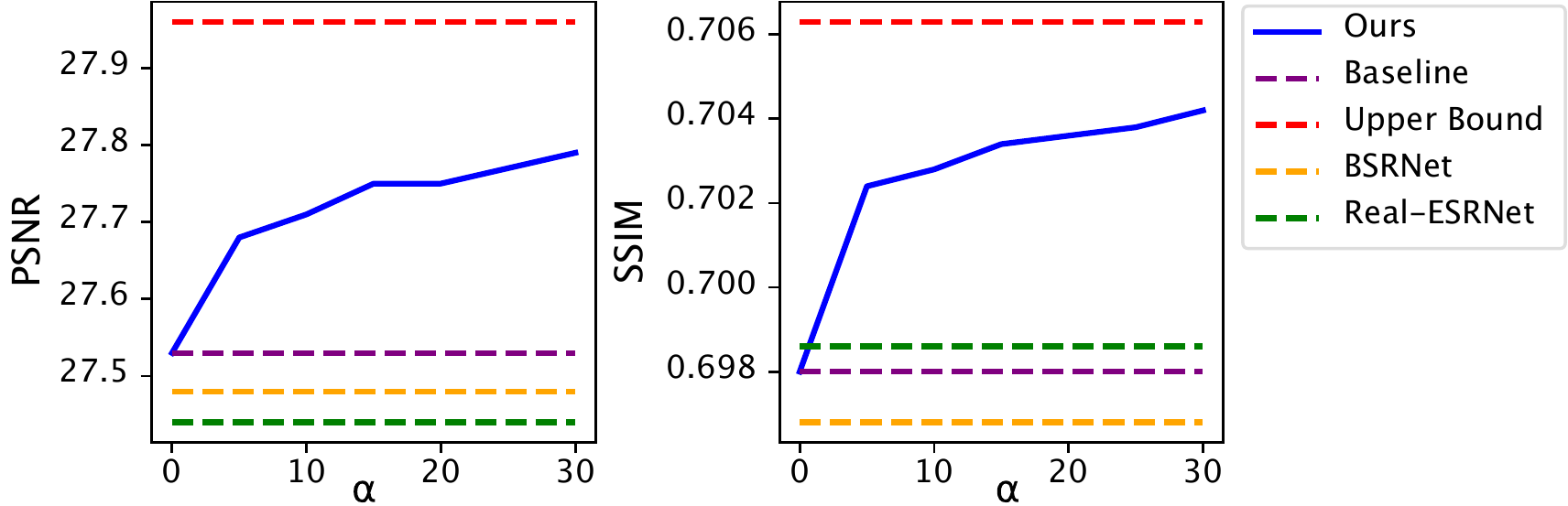}
   \vspace{-8mm}
   \caption{Average PSNR and SSIM performance at different $\alpha$s. This experiment uses Setting \ding{174} of \tablename~\ref{tab:synthetic_setting} as the target degradation distribution. The network structure is SRResNet. Its feature extractor is DASR.}
    \label{fig:alpha_change}
    \vspace{-6mm}
\end{figure}

\begin{table}[t]
    \centering
    \caption{The settings of the testing degradation distribution involved in our study.}
    \label{tab:synthetic_setting}
    \vspace{2mm}
    \resizebox{\linewidth}{!}{
    \begin{tabular}{c|c|c|c|c}
    \toprule
        Degradation & Setting \ding{172} & Setting \ding{173} & Setting \ding{174} & Setting \ding{175} \\
         \midrule
         Blur $\sigma$ & $\mathcal{U}_{[0,1]}$ & $\mathcal{U}_{[0.5,1.5]}$ & $\mathcal{U}_{[1.5,2.5]}$ & $\mathcal{U}_{[2.5,3.5]}$ \\
         Noise level $l$ & $\mathcal{U}_{[0,10]}$ & $\mathcal{U}_{[15,25]}$ & $\mathcal{U}_{[5,15]}$ & $\mathcal{U}_{[25,35]}$ \\
         JPEG quality $q$ & $\mathcal{U}_{[80,90]}$ & $\mathcal{U}_{[75,85]}$ & $\mathcal{U}_{[75,85]}$ & $\mathcal{U}_{[65,75]}$ \\
         \bottomrule
    \end{tabular}}
    \vspace{-5mm}
\end{table}

\begin{table*}
    \centering
    \caption{The quantitative comparison of different methods with respect to different settings. The upper bound results are marked as grey color to show that a direct comparison of it is unfair to other methods.}
    \label{tab:quant}
    \vspace{2mm}
    \resizebox{1.0\textwidth}{!}{
    \begin{tabular}{l|c|ccc|ccc|ccc|ccc}
    \toprule
        \multirow{2}{*}{Methods} & \multirow{2}{*}{Type} & \multicolumn{3}{c|}{Setting \ding{172}} & \multicolumn{3}{c|}{Setting \ding{173}} & \multicolumn{3}{c|}{Setting \ding{174}} & \multicolumn{3}{c}{Setting \ding{175}} \\
          & & PSNR & SSIM & LPIPS & PSNR & SSIM & LPIPS & PSNR & SSIM & LPIPS & PSNR & SSIM & LPIPS \\
         \midrule
         RCAN & non-blind & 23.65 & 0.5612 & 0.4674 & 18.86 & 0.2700 & 0.6509 & 24.22 & 0.5186 & 0.5812 & 16.54 & 0.1450 & 0.7152\\
         KernelGAN & blind & 16.04 & 0.3303 & 0.7420 & 12.56 & 0.1268 & 0.7675 & 17.79 & 0.3048 & 0.8097 & 11.34 & 0.0670 & 0.7828\\
         IKC & blind & 24.76 & 0.6174 & 0.4554 & 20.75 & 0.3606 & 0.6235 & 24.89 & 0.5607 & 0.5867 & 18.87 & 0.2323 & 0.6807\\
         DASR & blind & 24.81 & 0.6195 & 0.4505 & 20.11 & 0.3312 & 0.6432 & 24.75 & 0.5557 & 0.5938 & 17.64 & 0.1855 & 0.7103\\
         FSSR & blind & 20.83 & 0.4166 & 0.4194 & 16.14 & 0.1521 & 0.6989 & 20.75 & 0.3338 & 0.5189 & 14.75 & 0.0872 & 0.7686\\
         USR-DA & reference & 24.98 & 0.6712 & 0.3563 & 22.41 & 0.4816 & 0.5154 & 23.50 & 0.6076 & 0.4922 & 20.73 & 0.3428 & 0.6120\\
         PDM-SR & reference & 26.96 & 0.7071 & 0.3249 & 24.93 & 0.6208 & 0.4251 & 25.78 & 0.6433 & 0.4033 & 23.47 & 0.5576 & 0.4808\\
         BSRNet & blind & 27.88 & 0.7252 & 0.3431 & 26.80 & 0.6908 & 0.3964 & 27.48 & 0.6968 & 0.3987 & 25.01 & 0.6269 & 0.4916\\
         Real-ESRNet & blind & 27.75 & 0.7316 & 0.3410 & 26.75 & 0.6955 & 0.3896 & 27.44 & 0.6986 & 0.3994 & 25.36 & 0.6309 & 0.4916\\
         Real-ESRNet-Dropout & blind & 26.94 & 0.7123 & 0.3783 & 26.40 & 0.6777 & 0.4125 & 26.88 & 0.6879 & 0.4207 & 24.84 & 0.6109 & 0.5040 \\
         \midrule
         SRResNet (baseline) & blind & 28.14 & 0.7368 & 0.3370 & 27.01 & 0.6950 & 0.3873 & 27.53 & 0.6980 & 0.4067 & 25.34 & 0.6276 & 0.4992\\
         Ours ($\phi=$VGG$_{2,2}$) & reference & 28.85 & 0.7508 & 0.3173 & 27.32 & 0.7021 & 0.3861 & 27.77 & 0.7039 & 0.4011 & 25.50 & 0.6315 & 0.4952\\
         Ours ($\phi=$VGG$_{5,4}$) & reference & 28.82 & 0.7492 & 0.3192 & 27.36 & 0.7034 & 0.3824 & 27.73 & 0.7037 & 0.4005 & 25.43 & 0.6310 & 0.4950\\
         Ours ($\phi=$DASR) & reference & 28.88 & 0.7518 & 0.3147 & 27.27 & 0.7007 & 0.3866 & 27.79 & 0.7041 & 0.4005 & 25.48 & 0.6307 & 0.4962\\
         Ours ($\phi=$AlexNet-ImageNet) & reference & 28.83 & 0.7511 & 0.3166 & 27.42 & 0.7040 & 0.3836 & 27.89 & 0.7054 & 0.4002 & 25.21 & 0.6158 & 0.4871\\
         Ours ($\phi=$AlexNet-MINC) & reference & 28.85 & 0.7502 & 0.3193 & 27.44 & 0.7042 & 0.3815 & 27.99 & 0.7051 & 0.4048 & 25.53 & 0.6314 & 0.4959\\
         Ours ($\phi=$AlexNet-random) & reference & 28.77 & 0.7467 & 0.3256 & 27.44 & 0.7042 & 0.3815 & 27.85 & 0.7044 & 0.4020 & 25.54 & 0.6316 & 0.4957\\
         % Ours ($\phi=$SRResNet) & - & - & - & - & - & - & - & - & - & - & - & -\\
         \textcolor{gray}{Upper bound} & supervised & \textcolor{gray}{28.86} & \textcolor{gray}{0.7529} & \textcolor{gray}{0.3099} & \textcolor{gray}{27.44} & \textcolor{gray}{0.7054} & \textcolor{gray}{0.3792} & \textcolor{gray}{27.97} & \textcolor{gray}{0.7067} & \textcolor{gray}{0.4002} & \textcolor{gray}{25.59} & \textcolor{gray}{0.6321} & \textcolor{gray}{0.4975}\\
         \midrule
         RRDB (baseline) & blind & 28.70 & 0.7532 & 0.3035 & 27.45 & 0.7090 & 0.3648 & 27.91 & 0.7096 & 0.3875 & 25.63 & 0.6358 & 0.4881\\
         Ours ($\phi=$VGG$_{2,2}$) & reference & 29.18 & 0.7615 & 0.2970 & 27.66 & 0.7128 & 0.3658 & 28.07 & 0.7123 & 0.3863 & 25.74 & 0.6381 & 0.4826\\
         Ours ($\phi=$VGG$_{5,4}$) & reference & 29.16 & 0.7610 & 0.2973 & 27.73 & 0.7141 & 0.3640 & 28.07 & 0.7126 & 0.3856 & 25.68 & 0.6375 & 0.4824\\
         Ours ($\phi=$DASR) & reference & 29.18 & 0.7621 & 0.2957 & 27.65 & 0.7125 & 0.3663 & 28.08 & 0.7123 & 0.3861 & 25.73 & 0.6377 & 0.4832\\
         Ours ($\phi=$AlexNet-ImageNet) & reference & 29.16 & 0.7614 & 0.2963 & 27.77 & 0.7142 & 0.3649 & 28.17 & 0.7132 & 0.3862 & 25.76 & 0.6378 & 0.4834\\
         Ours ($\phi=$AlexNet-MINC) & reference & 29.18 & 0.7613 & 0.2978 & 27.77 & 0.7142 & 0.3645 & 28.24 & 0.7133 & 0.3891 & 25.79 & 0.6381 & 0.4838\\
         \textcolor{gray}{Upper bound} & supervised & \textcolor{gray}{29.15} & \textcolor{gray}{0.7607} & \textcolor{gray}{0.2979} & \textcolor{gray}{27.75} & \textcolor{gray}{0.7135} & \textcolor{gray}{0.3680} & \textcolor{gray}{28.23} & \textcolor{gray}{0.7135} & \textcolor{gray}{0.3888} & \textcolor{gray}{25.84} & \textcolor{gray}{0.6384} & \textcolor{gray}{0.4892}\\
         \bottomrule
    \end{tabular}}
    \vspace{-3mm}
\end{table*}

\textbf{The choice of the deep feature extractor $\phi$.}
We use a deep network to extract features to assist in computing the Fréchet distance between degradation distributions.
But is this distance computation robust to the choice of feature extractor?
Do we need to train a deep feature extractor specifically for this distance?
In this subsection, we verify the effectiveness of the proposed degradation Fréchet distance and the difference between feature extractors.
We selected the following representative deep feature extractors: VGG$_{2,2}$\footnote{VGG$_{i,j}$ is defined as the feature map obtained by the $j$th convolution (after activation) and before the $i$th max-pooling layer within the VGG19 \cite{simonyan2014very} network} is a commonly used extractor for extracting low-level features; VGG$_{5,4}$, which is often used for extracting deeper features; the DASR degradation representation \cite{wang2021unsupervised}, which is specially used to learn the latent representation for degradation; AlexNet trained using ImageNet dataset \cite{deng2009imagenet}, MINC texture classification dataset \cite{bell2015material}; and a randomly initialized AlexNet.
The results are visualized in \figurename~\ref{fig:alpha}.
We arrange these 75 bins regularly and visualize their calculated sampling weights. Weights closer to ground truth are better.
It can be seen that these methods can identify the four bins that appear in the ground truth and give them higher weights. However, these feature extractors perform differently on the weight calculation of the remaining bins.
Both AlexNet and VGG trained on ImageNet can predict weights with better results. AlexNet trained with MINC can also achieve good results. These methods only assign minor weight to non-target degradation bins.
Although DASR is a method specially designed for degradation, its prediction assigns too much weight to irrelevant bins, which may affect its performance.
Randomly initialized AlexNet can only show limited effectiveness in matching degradation distributions.
In the following research, we mainly use AlexNet because of its good prediction effect and easy availability.

\textbf{The Use of the Fréchet distance.}
In this work, the Fréchet distance is used to overcome the influence of different image content on degenerate distance estimation.
In the case of image content changes, the traditional element-wise comparison is not applicable anymore.
However, the Fréchet distance can produce reasonable results.
The Fréchet distance is based on statistics on features, so the number of samples used to estimate this distance is important.
We made estimations using different numbers of images and studied the variance of different measurements.
We tested each image quantity 25 times with different images and recorded their values.
The size of the image is $72\times72$.
The results are shown in \figurename~\ref{fig:box_curve} (c).
It can be seen that when 100 images are used, the randomness of the results obtained by the algorithm is greatly reduced.
Although more images can bring better stability, 100 -- 150 images can already give good estimation results.
In contrast, the previous works that allow reference testing images are based on adversarial learning \cite{luo2022learning}, or domain adaptation matching \cite{wang2021domain}.
They usually require a large number of reference images and produce unsatisfactory results when the number of reference images is insufficient.

\textbf{The choice of the distribution range $\alpha$.}
Another important parameter in our method is $\alpha$, which scales the Fréchet distance by a power function to adjust the range of the final degenerate distribution.
$\alpha$ is the only parameter in our method that needs to be adjusted manually.
A smaller $\alpha$ means a larger degradation range and better generalization performance.
But smaller $\alpha$ also leads to lower SR performance.
A larger $\alpha$ means a narrower degradation range, which will improve the resulting final performance when the target degradation range is also small.
However, the performance degradation faced when exceeding this range is also more severe.
$\alpha$ controls the accuracy-generalization trade-off.
We show the visualized weights using different $\alpha$s in \figurename~\ref{fig:alpha}.
In order to verify the impact of different $\alpha$s on performance, we set a synthetic test degradation interval. And use different $\alpha$s to obtain the training degradation weights.
The baseline is the result of uniform sampling among the bins. The upper bound is the result of direct training on target degradation. %
As can be seen, as $\alpha$ increases, the range of training degradation becomes smaller, and the test accuracy within the target range is improved.
We also include Real-ESRNet and BSRNet for comparison.
Their training degradation range is larger than the baseline, so even though they use a better network, their only achieve lower performance than the proposed method.

\begin{figure*}[t]
%\newlength-4mm
%\setlength{-4mm}{-0.4cm}
\scriptsize
\centering
\resizebox{1\textwidth}{!}{
\begin{tabular}{ccc}

% % % % one row

\hspace{-0.45cm}
\begin{adjustbox}{valign=t}
\begin{tabular}{c}
\includegraphics[width=0.211\textwidth, viewport=0  0 470 575, clip]{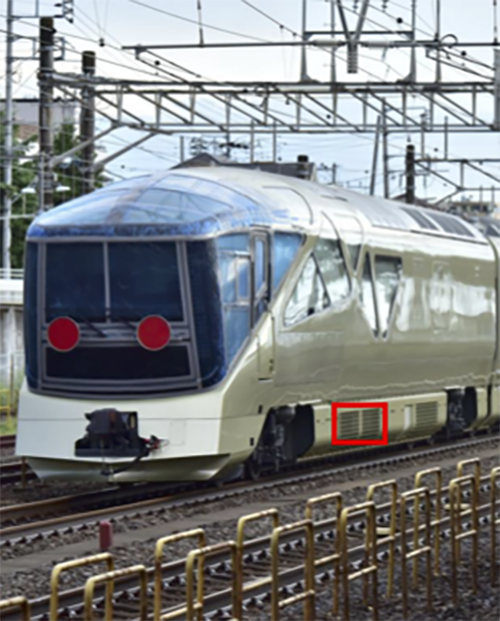}
\\
% 0820
DIV2K 0820, degradation \ding{172}
\end{tabular}
\end{adjustbox}
\hspace{-0.46cm}
\begin{adjustbox}{valign=t}
\begin{tabular}{cccccc}
\includegraphics[width=0.18\textwidth]{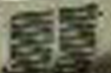} \hspace{-4mm} &
\includegraphics[width=0.18\textwidth]{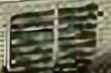} \hspace{-4mm} &
\includegraphics[width=0.18\textwidth]{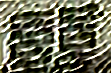} \hspace{-4mm} &
\includegraphics[width=0.18\textwidth]{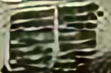} \hspace{-4mm} &
\includegraphics[width=0.18\textwidth]{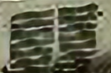} \hspace{-4mm} &
\includegraphics[width=0.18\textwidth]{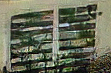} \hspace{-4mm} 
\\
Bicubic  \hspace{-4mm} &
RCAN  \hspace{-4mm} &
KernelGAN  \hspace{-4mm} &
IKC  \hspace{-4mm} &
DASR  \hspace{-4mm} &
FSSR  \hspace{-4mm}
\\
\includegraphics[width=0.18\textwidth]{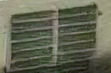} \hspace{-4mm} &
\includegraphics[width=0.18\textwidth]{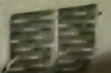} \hspace{-4mm} &
\includegraphics[width=0.18\textwidth]{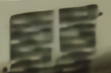} \hspace{-4mm} &
\includegraphics[width=0.18\textwidth]{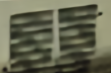} \hspace{-4mm} &
\includegraphics[width=0.18\textwidth]{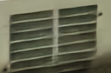} \hspace{-4mm} &
\includegraphics[width=0.18\textwidth]{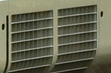} \hspace{-4mm}  
\\ 
USR-DA   \hspace{-4mm} &
PDM-SR  \hspace{-4mm} &
BSRNet  \hspace{-4mm} &
Real-ESRNet  \hspace{-4mm} &
Ours \hspace{-4mm} &
GT \hspace{-4mm}
\\
\end{tabular}
\end{adjustbox}
\\
% % % % one row
\end{tabular}}
\vspace{-4mm}
\caption{SR results of images from the DIV2K dataset with scale factor $\times$4. 
% The testing image is synthesized by degradation setting \ding{172}.
}
\label{fig:synsr}
\vspace{-2mm}
\end{figure*}

% =================================================================

\begin{figure*}[t]
%\newlength-4mm
%\setlength{-4mm}{-0.4cm}
\scriptsize
\centering
\resizebox{1\textwidth}{!}{
\begin{tabular}{ccc}
% % one row
\hspace{-0.45cm}
\begin{adjustbox}{valign=t}
\begin{tabular}{c}
\includegraphics[width=0.211\textwidth, viewport=0  0 2000 1630, clip]{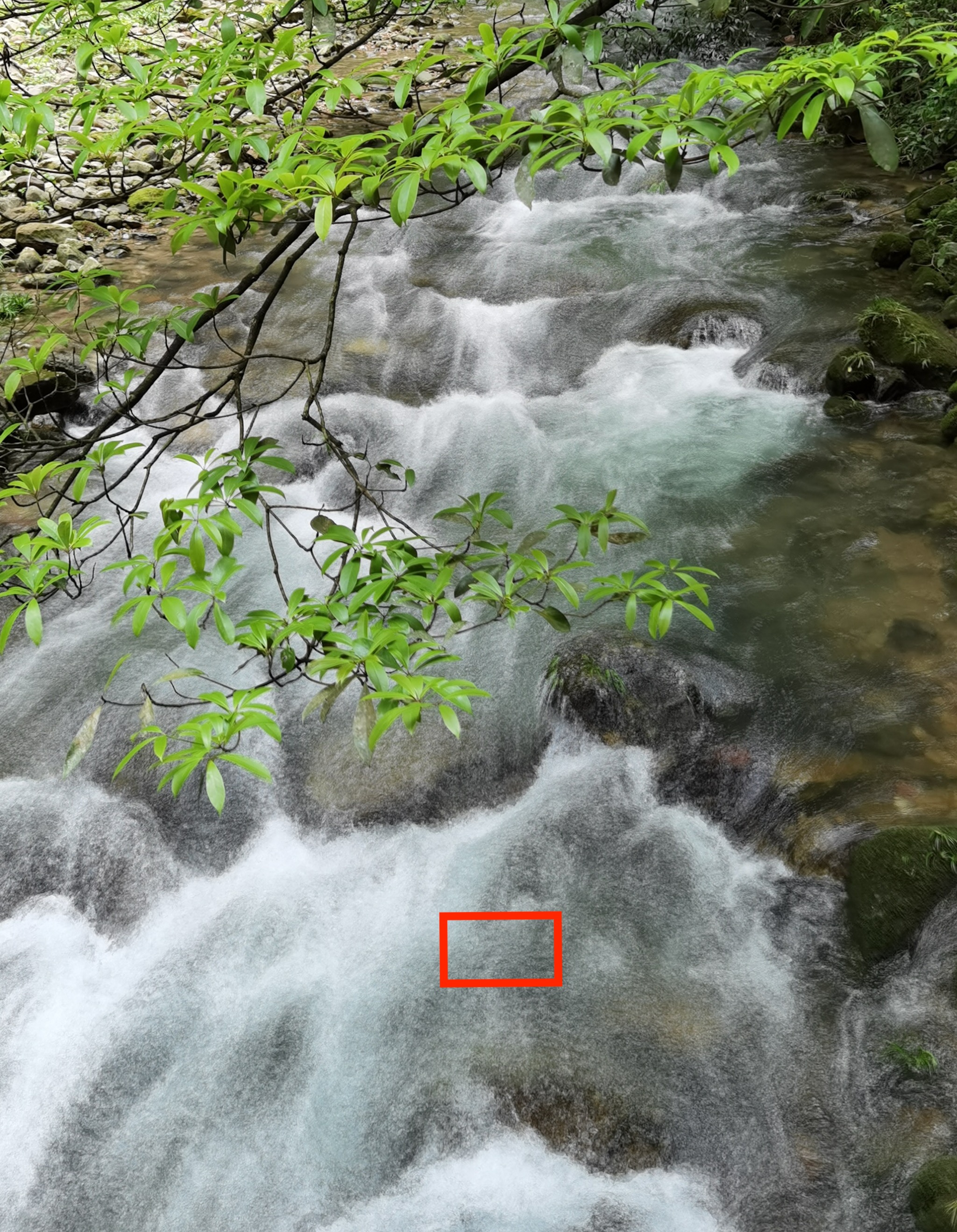}
\\
% P40 IMG\_02
% Case 1
HUAWEI P40
\end{tabular}
\end{adjustbox}
\hspace{-0.46cm}
\begin{adjustbox}{valign=t}
\begin{tabular}{cccccc}
\includegraphics[width=0.18\textwidth, viewport=0  0 1200 500, clip]{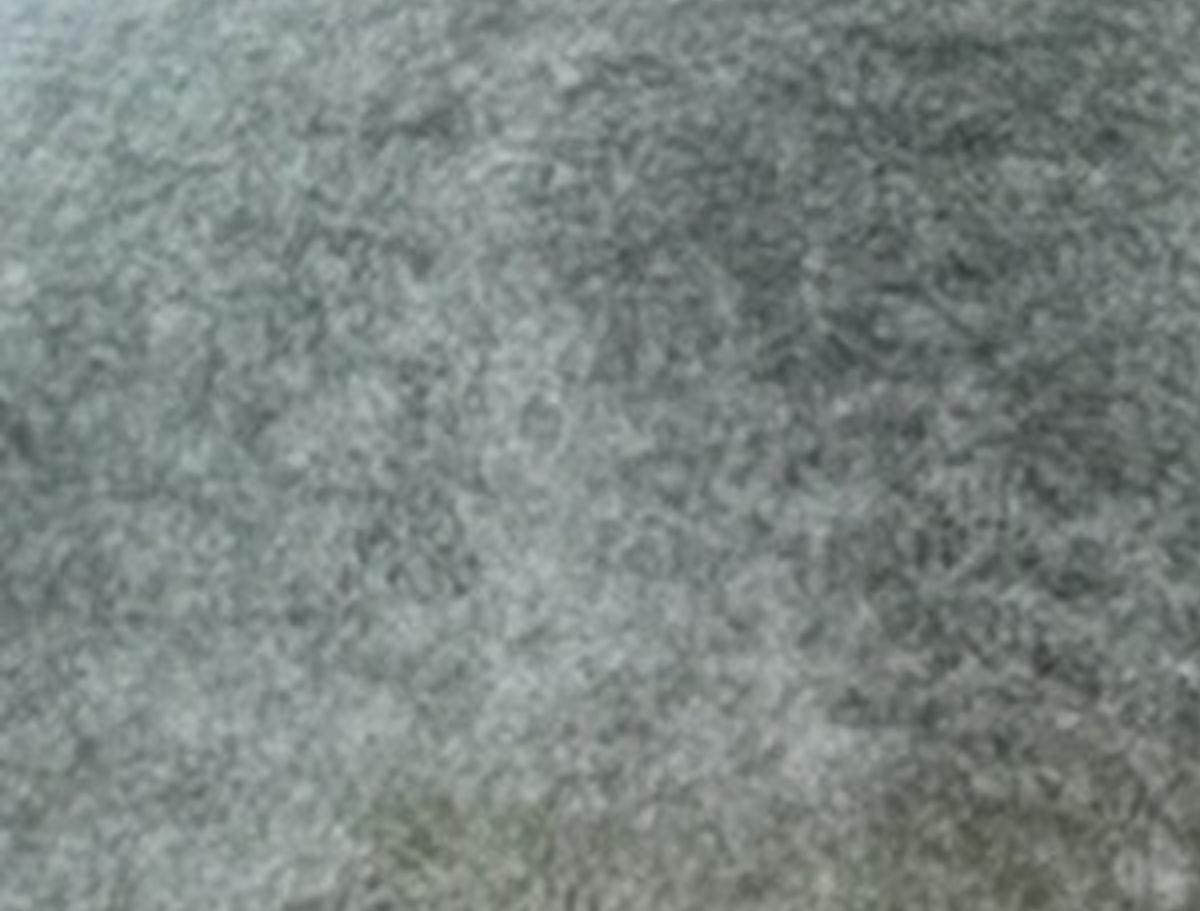} \hspace{-4mm} &
\includegraphics[width=0.18\textwidth, viewport=0  0 1200 500, clip]{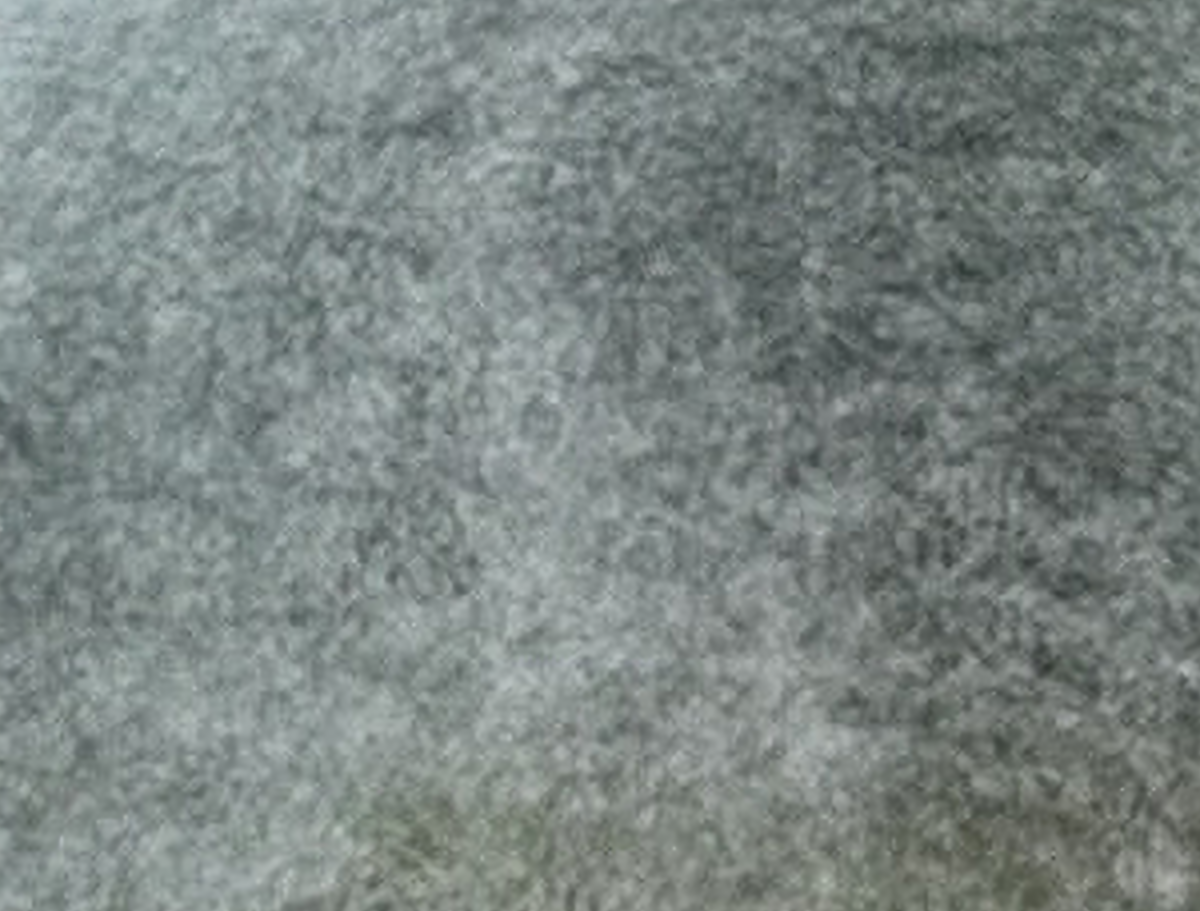} \hspace{-4mm} &
\includegraphics[width=0.18\textwidth, viewport=0  0 1200 500, clip]{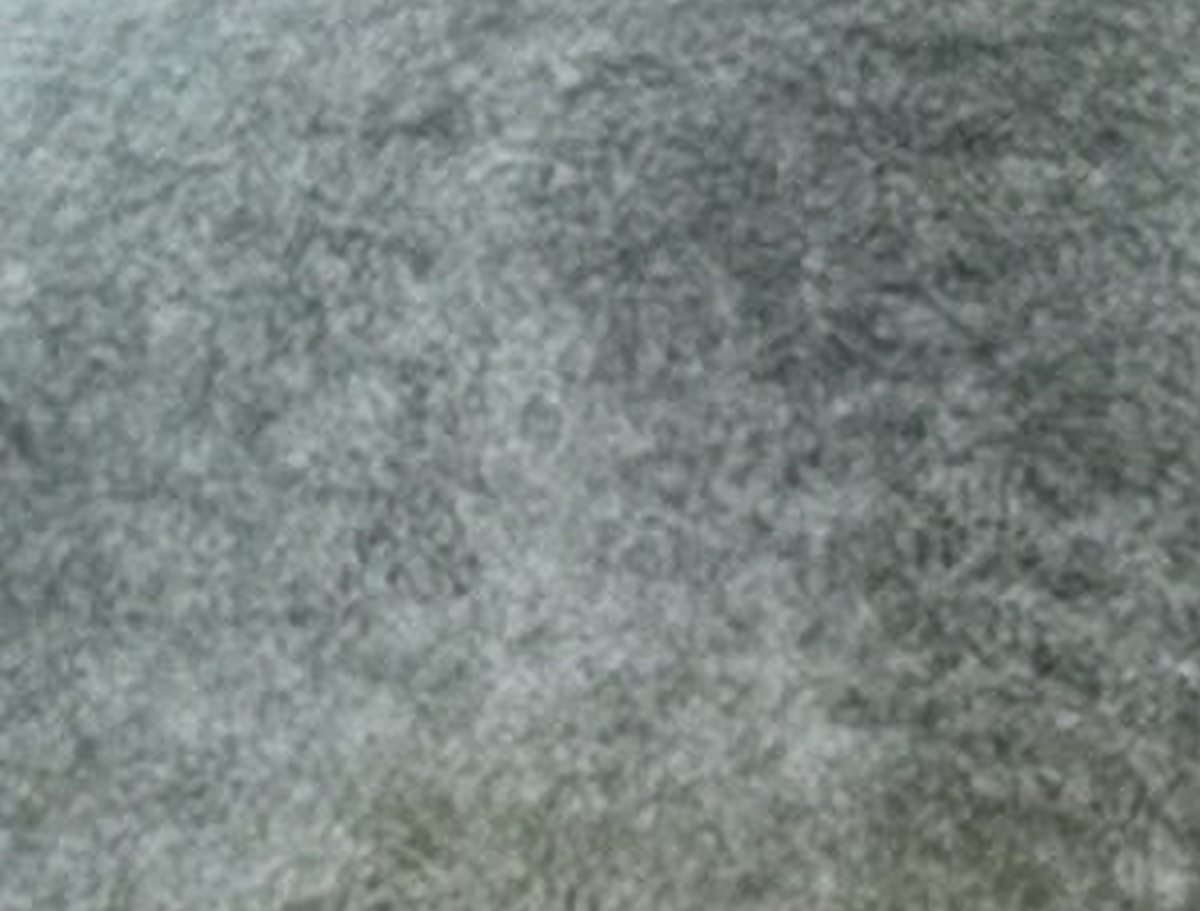} \hspace{-4mm} &
\includegraphics[width=0.18\textwidth, viewport=0  0 1200 500, clip]{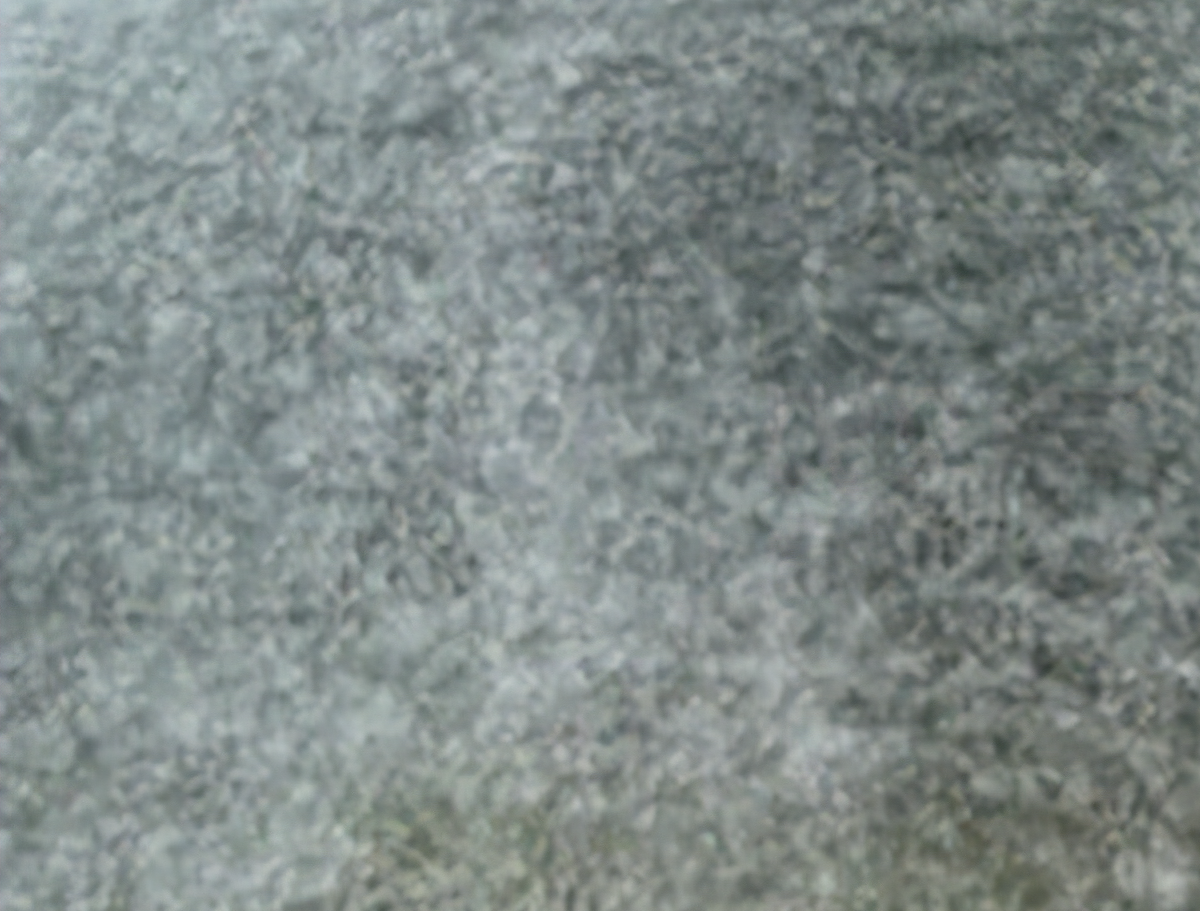} \hspace{-4mm} 
\\
Bicubic  \hspace{-4mm} &
IKC  \hspace{-4mm} &
DASR  \hspace{-4mm} &
FSSR  \hspace{-4mm}
\\
\includegraphics[width=0.18\textwidth, viewport=0  0 1200 500, clip]{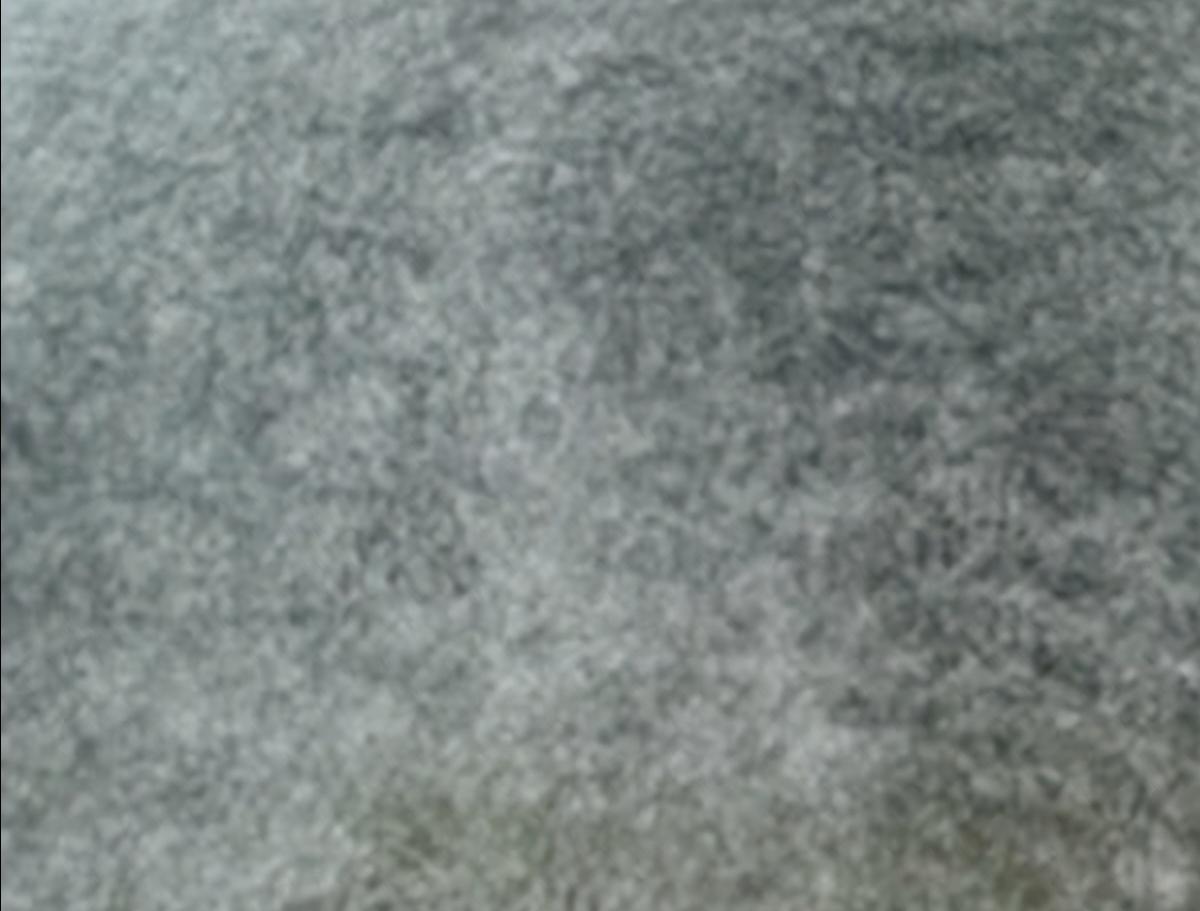} \hspace{-4mm} &
\includegraphics[width=0.18\textwidth, viewport=0  0 1200 500, clip]{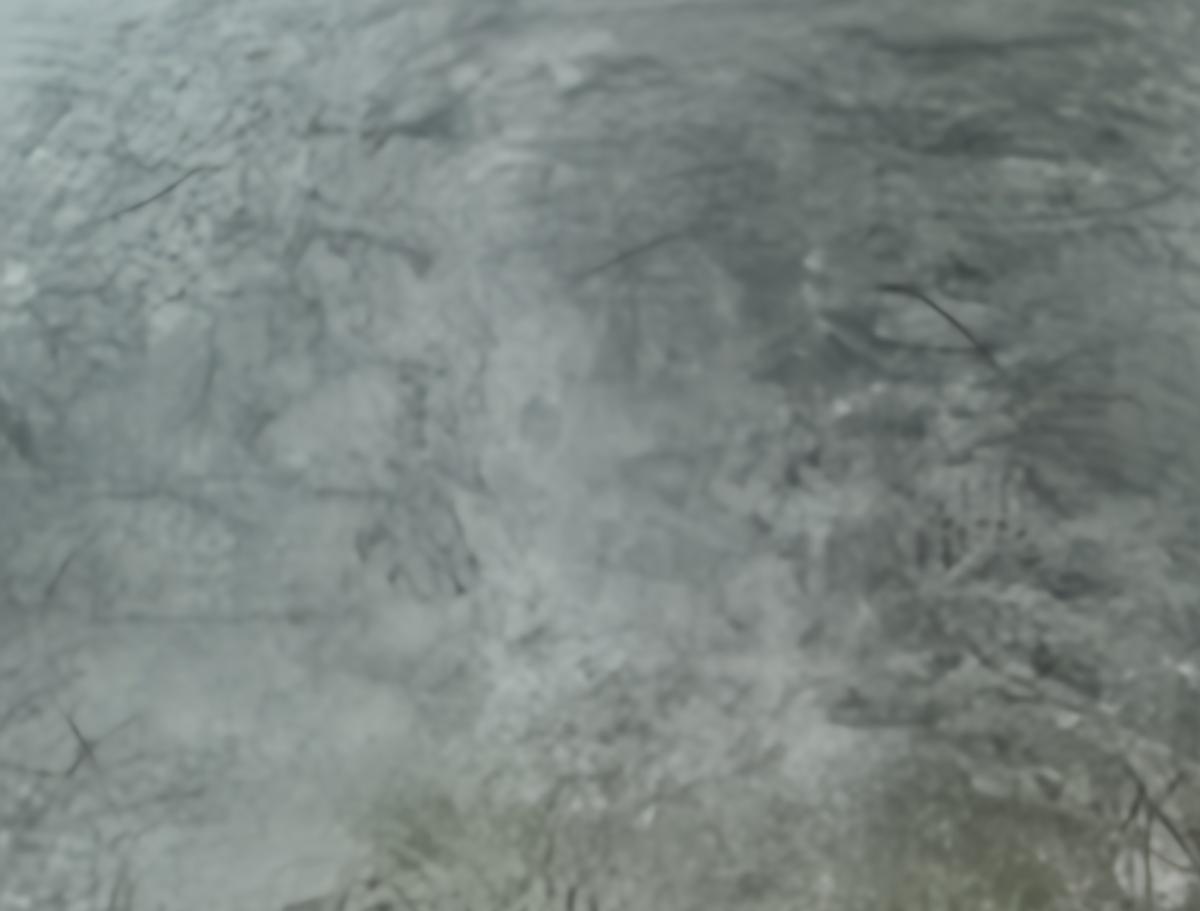} \hspace{-4mm} &
\includegraphics[width=0.18\textwidth, viewport=0  0 1200 500, clip]{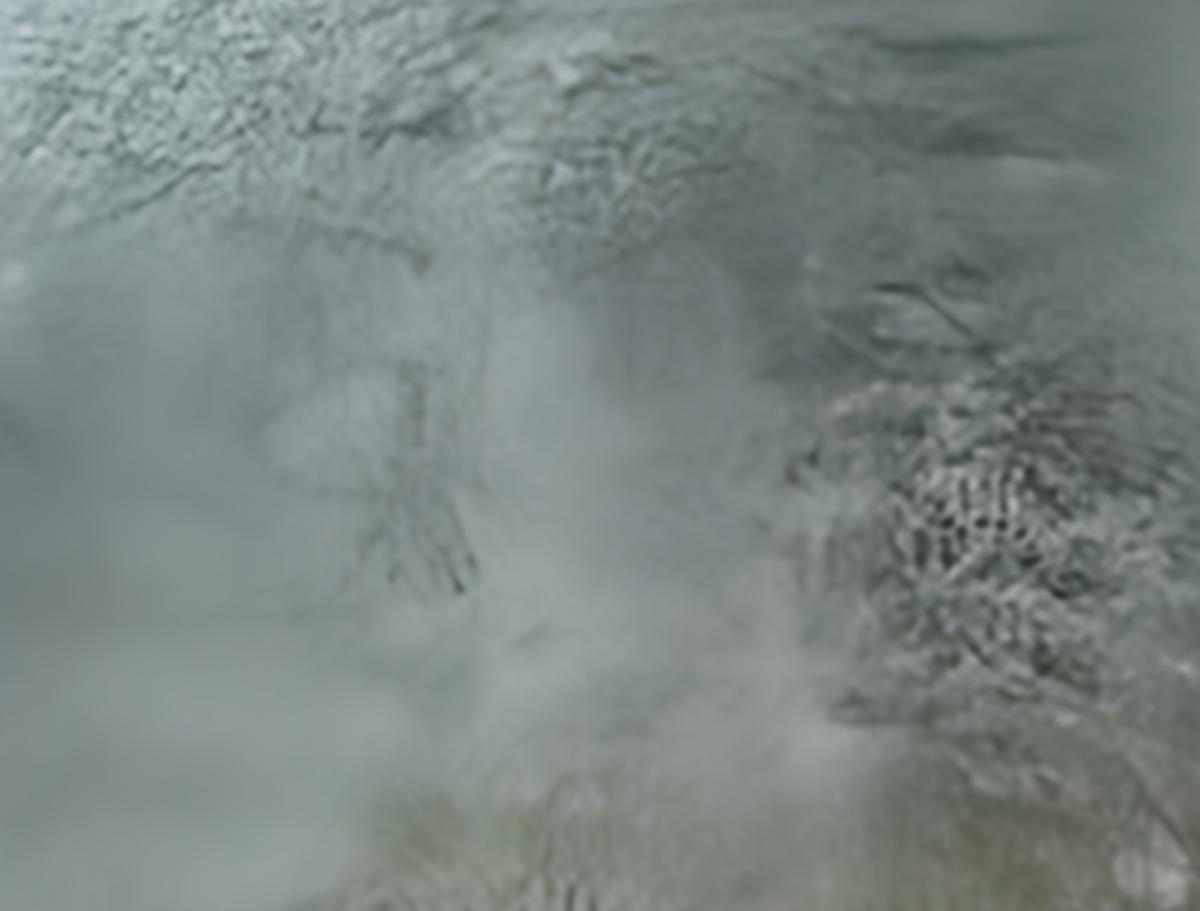} \hspace{-4mm} &
\includegraphics[width=0.18\textwidth, viewport=0  0 1200 500, clip]{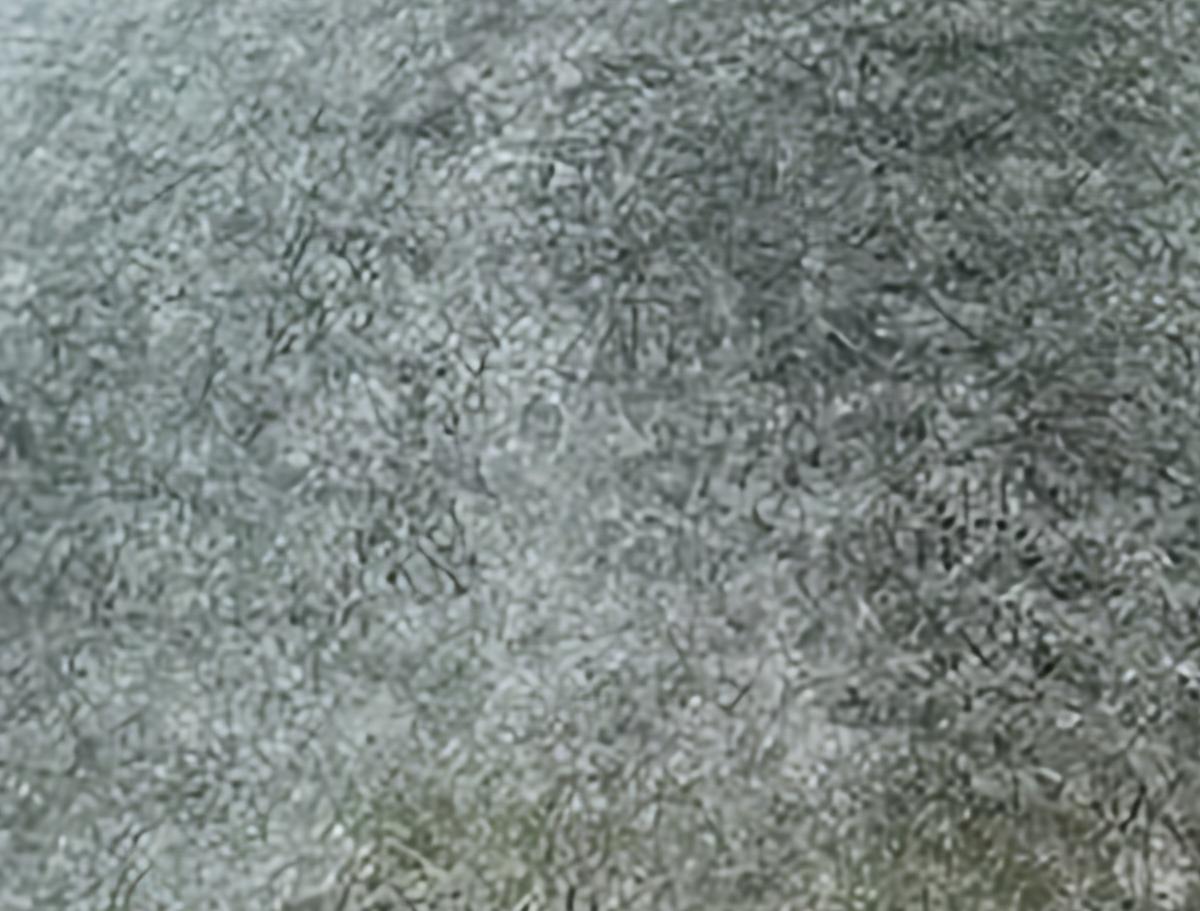} \hspace{-4mm}  
\\ 
USR-DA   \hspace{-4mm} &
BSRNet  \hspace{-4mm} &
Real-ESRNet \hspace{-4mm} &
Ours \hspace{-4mm}
\\
\end{tabular}
\end{adjustbox}
\\

\hspace{-0.45cm}
\begin{adjustbox}{valign=t}
\begin{tabular}{c}
\includegraphics[width=0.211\textwidth, viewport=135  0 500 400, clip]{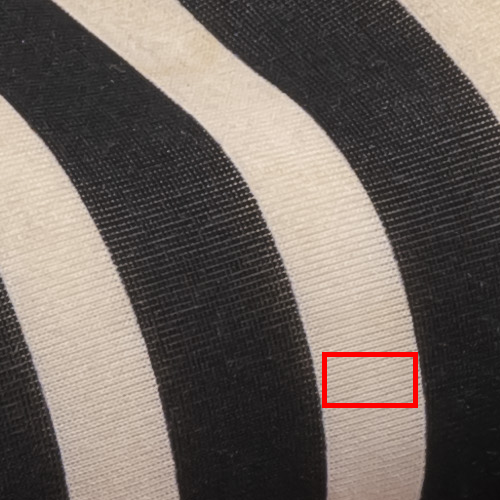}
\\
Nikon
\end{tabular}
\end{adjustbox}
\hspace{-0.46cm}
\begin{adjustbox}{valign=t}
\begin{tabular}{cccccc}
\includegraphics[width=0.18\textwidth]{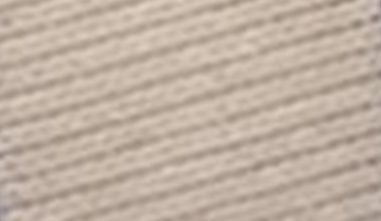} \hspace{-4mm} &
\includegraphics[width=0.18\textwidth]{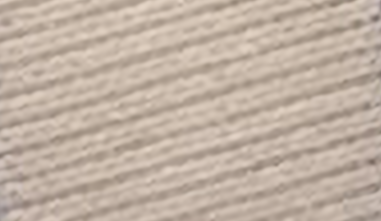} \hspace{-4mm} &
\includegraphics[width=0.18\textwidth]{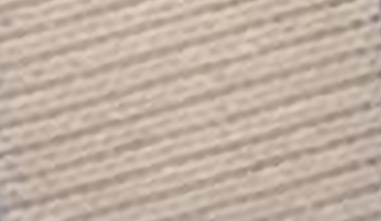} \hspace{-4mm} &
\includegraphics[width=0.18\textwidth]{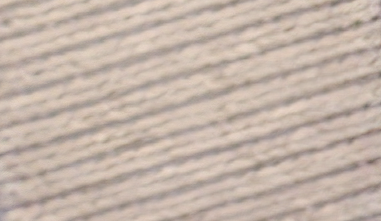} \hspace{-4mm} 
\\
Bicubic  \hspace{-4mm} &
IKC  \hspace{-4mm} &
DASR  \hspace{-4mm} &
FSSR  \hspace{-4mm}
\\
\includegraphics[width=0.18\textwidth]{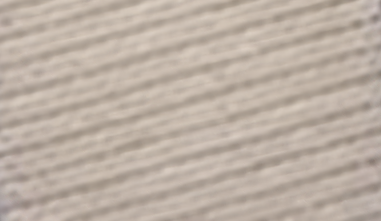} \hspace{-4mm} &
\includegraphics[width=0.18\textwidth]{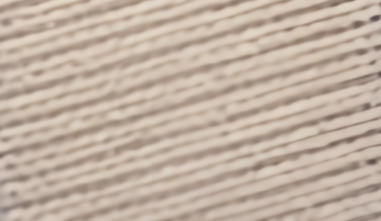} \hspace{-4mm} &
\includegraphics[width=0.18\textwidth]{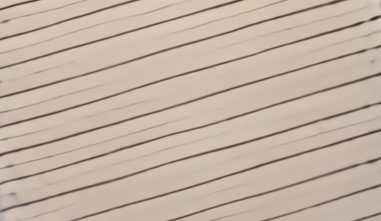} \hspace{-4mm} &
\includegraphics[width=0.18\textwidth]{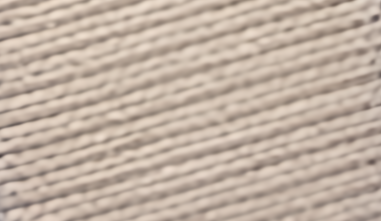} \hspace{-4mm}  
\\ 
USR-DA   \hspace{-4mm} &
BSRNet  \hspace{-4mm} &
Real-ESRNet \hspace{-4mm} &
Ours \hspace{-4mm}
\\
\end{tabular}
\end{adjustbox}
\\

% % % % one row
\end{tabular}}
\vspace{-4mm}
\caption{SR results of real cases with scale factor $\times$4. The first case is from a smartphone camera. The rest two cases are from old films.}
\label{fig:realsr}
\vspace{-3mm}
\end{figure*}

\vspace{-3mm}
\section{Experiments}

\subsection{Experiments on the Synthesized Images}
% \vspace{-2mm}

\paragraph{Set up.}
We first evaluate the performance of the proposed method on the synthetic test images.
We set four different cases with four different degradation distributions.
The detail of these settings is shown in \tablename~\ref{tab:synthetic_setting}.
We carefully designed these settings to include relatively clean test images (setting \ding{172} contains small blur and noise); two cases of moderate degeneration (setting \ding{173} contains smaller blur and larger noise and \ding{174} contains larger blur with smaller noise; and severely degeneration (setting \ding{175}).
We use the PIPAL dataset \cite{jinjin2020pipal} as the reference set $Y_{ref}$ and $Y_w$.
Following the previous study on the number of images used to calculate the Fréchet distance, we set the number of reference images as $n=100$, and the size of the degraded images is $72\times72$.
Although the degradation is randomly sampled from a distribution, we fix the degraded images during the experiment to eliminate the effect of randomness.
In our experiments, we included a total of 75 degradation distribution bins. For the image blurring, we divide the kernel width into five equal parts between 0 and 5. For the noise, we divide the noise level into five equal parts between 0 and 50. For image compression, divide the quality level into three equal parts between 30 and 90.
We set $\alpha=25$.

\vspace{-3mm}
\paragraph{Results.}
We compare the proposed method with several existing methods, including a non-blind method RCAN \cite{rcan2018}, two blind SR method IKC \cite{Gu_2019_CVPR} and KernelGAN \cite{bell2019blind}, DASR with a pre-defined degradation model, the FSSR model trained to maximize the performance on the blurry and noisy dataset, the BSRNet and Real-ESRNet trained with a large range of complex degradations and the Dropout method that proposed to improve generalization performance.
We also include two methods with reference images as input, USR-DA \cite{wang2021domain}, and PDM-SR \cite{luo2022learning}.
Additionally, we also compared the following methods: (1) the baseline model with a uniform sampling from all bins, (2) models with different feature extractors, and (3) the model trained with corresponding test degradation for each setting, which is used to show the upper bound performance in each setting.
We test two different backbone architectures: SRResNet and RRDB \cite{wang2018esrgan}.
The results are shown in \tablename~\ref{tab:quant}.
It can be seen that, despite the careful design, existing blind problems do not perform well under these stochastic degradations.
The reasons include the two aspects mentioned earlier: insufficient generalization for cases beyond the pre-defined degradation models degradation, and the use of a too-large training degradation distribution which leads to a drop in overall accuracy.
As for our method, we first provide a comparison of the performance of the baseline model with that of our method.
We can see that our method has a substantial performance improvement over the baseline without the degradation range crafting.
This is more evident for the clean image test dataset (the setting \ding{172}), as training with large degradation is unnecessary for these situations.
Methods using different feature extraction all provide good performance improvement compared to the baseline model and other competitive methods, which shows that our method is easy to use.
AlexNet trained on ImageNet and MINC outperform others, which is in line with the conclusion in \figurename~\ref{fig:alpha}.

The qualitative results agree with the conclusions of the numerical results.
We show a set of comparisons in the \figurename~\ref{fig:synsr}.
It can be seen that some blind methods are almost completely ineffective in this case, such as IKC, KernelGAN and FSSR.
Because the degradation models assumed by these methods' design do not match the test degradation.
BSRNet and Real-ESRNet have good generalization ability due to training on large-scale and complex degraded datasets, thus obtaining smooth and clear results. However, the training degradation range used by these methods is too large. 
Degraded training data outside the target range will affect the training, making them unable to generate accurate texture details.
Based on the proposed method, our method correctly recovers the pattern and produces a pleasing result. 
Please refer to the appendix for more results.

\begin{figure}[t]
    \centering
   \includegraphics[width=\linewidth]{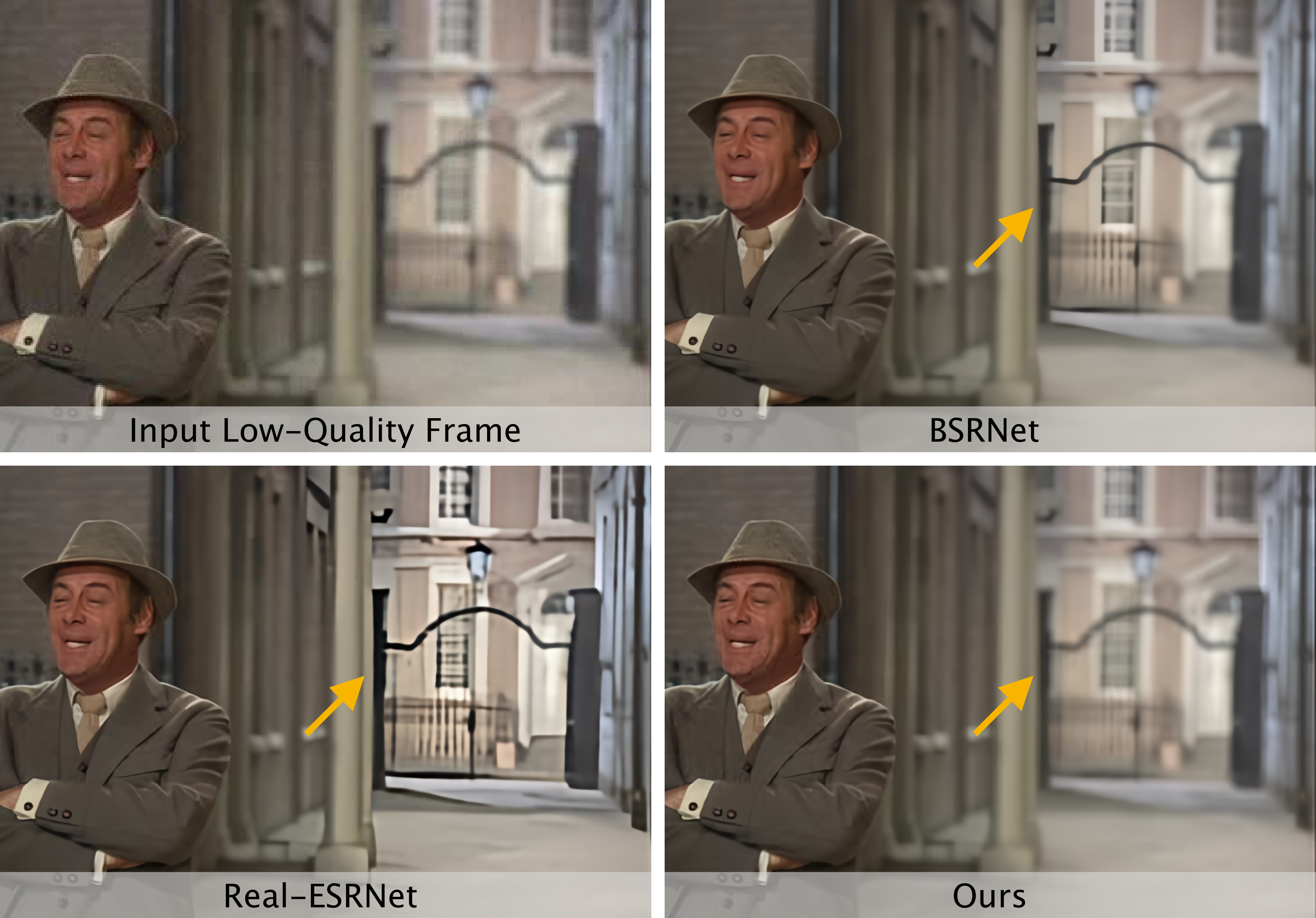}
   \vspace{-7mm}
   \caption{Due to the large range of training blur, Real-ESRNet and BSRNet may unreasonably sharpen the background blur. However, our method does not have this problem.}
    \label{fig:frames}
    \vspace{-6mm}
\end{figure}

\vspace{-2mm}
\subsection{Experiments for Real-World Applications}
\vspace{-2mm}
Our method is designed for real-world scenarios.
We demonstrate the value of our research using two valuable scenarios: SR processing of images taken by DSLR (digital single-lens reflex) cameras or mobile phones and old films.
The processing of these images is an admittedly difficult task due to the complexities of noise and blur.
But we argue that, even though complex, these degradations are not as wide-ranging as we thought.
Especially when we limit our processing to only a certain class of sensors or lenses.
The range of this test degradation will be further narrowed.
This is very suitable for the method proposed in this paper.
We test our method on the RealSR dataset \cite{cai2019toward} and our collected mobile phone camera dataset.
RealSR contains images taken by Canon 5D3 DSLR cameras.
The mobile data is captured by Huawei P40 mobile phone.
Furthermore, we also extracted the image frames of two old films as the testing sets. These two films are ``Groundhog Day'' and ``My Fair Lady''.
We show some visual results in \figurename~\ref{fig:realsr}.

As one can see, our results show excellent sharpness and detail restoration.
Some methods, such as IKC and DASR, lead to ambiguous results due to the mismatch of their degradation models.
An example is the image from ``Groundhog Day''  in \figurename~\ref{fig:realsr}. The results obtained by these two methods are still blurry, while our method is able to recover the sharp edges.
BSRNet and Real-ESRNet can handle a wide range of degradations at the cost of reduced accuracy. However, due to the larger range of noise and blur used for training, the network tends to reduce all noise-like textures. This makes it unable to recover some subtle textures and generate over-smooth results.
This drawback is evident in the case of the image captured by Huawei P40  in \figurename~\ref{fig:realsr}. A large area of the dense texture is removed by BSRNet and Real-ESRNet, resulting in over-smooth results, while our method can restore sufficient texture.
For the testing image from ``My Fair Lady'', since with closer training degradation distribution, only our method recovers the correct pattern of the image, other methods all result in incorrect patterns, especially USR-DA and BSRNet, which directly recover the circular pattern into lines.
Another problem with the existing methods is the processing of background blur.
As an artistic technique, background blur often appears on film screens.
Due to the large range of training blur, Real-ESRNet and BSRNet may unreasonably sharpen the background blur, as shown in \figurename~\ref{fig:frames}. However, our method does not have this problem.
These results demonstrate that our method achieves pleasing edges and effects while preserving detail, and also matches the correct pattern of the image.

\begin{figure}[t]
    \centering
   \includegraphics[width=\linewidth]{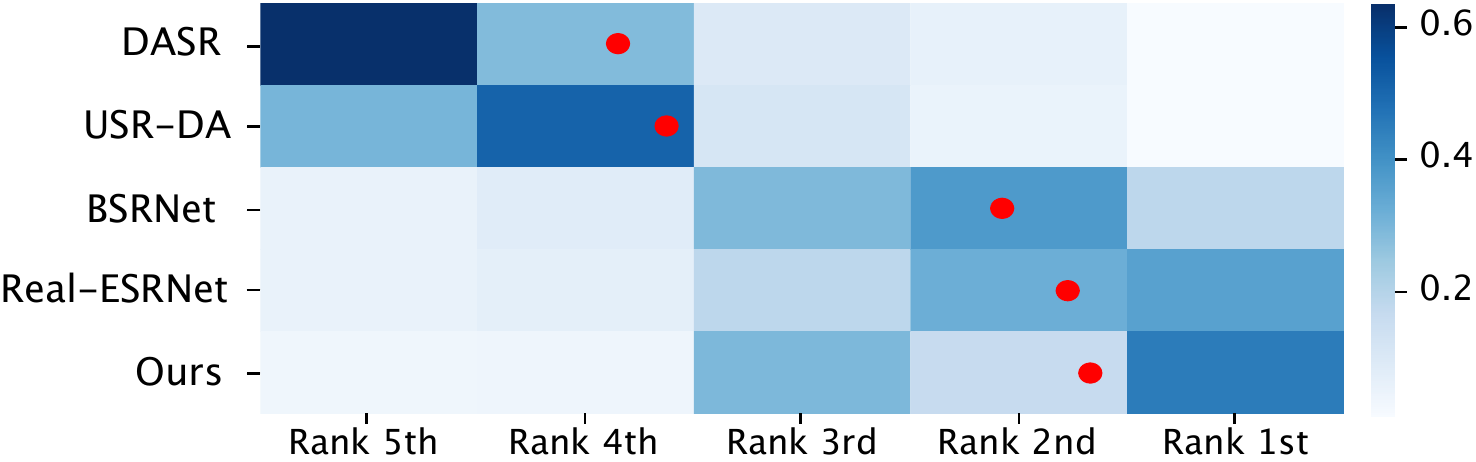}
   \vspace{-7mm}
   \caption{This figure shows the normalized histogram of votes in the user study. The average score is also shown in red dots.}
    \label{fig:userstudy}
    \vspace{-6mm}
\end{figure}

Due to the lack of reasonable quantitative measures for comparing real images. We conducted a user study for some representative methods.
We show the results of the five methods and ask the user to rank them. In total, our research contains more than 20 images from different scenes and sources. More than 30 users participated in our user study shown in \figurename~\ref{fig:userstudy}.
Our method was ranked first most times, and its average score also outperformed other methods.

\vspace{-2mm}
\section{Conclusion and Limitation}
\vspace{-2mm}
This paper describes a method to craft the training degradation distribution for real-world SR applications. Our method can improve its performance while maintaining the generalization ability of SR.
One of the limitations of our work is that the number of bins increases exponentially as the degradation model becomes more complex.

\vspace{-3mm}
\paragraph*{Acknowledgements}
This work was partly supported by the National Natural Science Foundation of China (Nos. 62276251, 62171251 and 62311530100), the Special Foundations for the Development of Strategic Emerging Industries of Shenzhen (Nos.JCYJ20200109143010272 and CJGJZD20210408092804011), Oversea Cooperation Foundation of Tsinghua, the Joint Lab of CAS-HK, and in part by the Youth Innovation Promotion Association of Chinese Academy of Sciences (No. 2020356).

\nocite{langley00}

\bibliography{example_paper}
\bibliographystyle{icml2023}

\newpage
\appendix
\onecolumn

\section{More Results}
We provide more qualitative results to show the effectiveness of our method.

\begin{figure*}[!hb]
%\newlength-4mm
%\setlength{-4mm}{-0.4cm}
\scriptsize
\centering
\resizebox{1\textwidth}{!}{
\begin{tabular}{ccc}
% % one row
\hspace{-0.45cm}
\begin{adjustbox}{valign=t}
\begin{tabular}{c}
\includegraphics[width=0.211\textwidth]{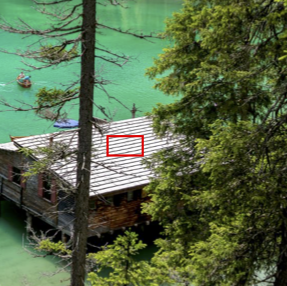}
\\
DIV2K 0807, degradation \ding{172}
\end{tabular}
\end{adjustbox}
\hspace{-0.46cm}
\begin{adjustbox}{valign=t}
\begin{tabular}{cccccc}
\includegraphics[width=0.18\textwidth]{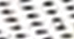} \hspace{-4mm} &
\includegraphics[width=0.18\textwidth]{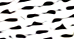} \hspace{-4mm} &
\includegraphics[width=0.18\textwidth]{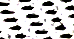} \hspace{-4mm} &
\includegraphics[width=0.18\textwidth]{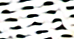} \hspace{-4mm} &
\includegraphics[width=0.18\textwidth]{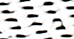} \hspace{-4mm} &
\includegraphics[width=0.18\textwidth]{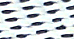} \hspace{-4mm} 
\\
Bicubic  \hspace{-4mm} &
RCAN  \hspace{-4mm} &
KernelGAN  \hspace{-4mm} &
IKC  \hspace{-4mm} &
DASR  \hspace{-4mm} &
FSSR  \hspace{-4mm}
\\
\includegraphics[width=0.18\textwidth]{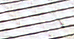} \hspace{-4mm} &
\includegraphics[width=0.18\textwidth]{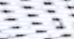} \hspace{-4mm} &
\includegraphics[width=0.18\textwidth]{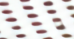} \hspace{-4mm} &
\includegraphics[width=0.18\textwidth]{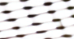} \hspace{-4mm} &
\includegraphics[width=0.18\textwidth]{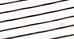} \hspace{-4mm} &
\includegraphics[width=0.18\textwidth]{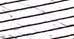} \hspace{-4mm}  
\\ 
USR-DA   \hspace{-4mm} &
PDM-SR  \hspace{-4mm} &
BSRNet  \hspace{-4mm} &
Real-ESRNet  \hspace{-4mm} &
Ours \hspace{-4mm} &
GT \hspace{-4mm}
\\
\end{tabular}
\end{adjustbox}
\\

% % % % one row

\hspace{-0.45cm}
\begin{adjustbox}{valign=t}
\begin{tabular}{c}
\includegraphics[width=0.211\textwidth, viewport=0  126 380 704, clip]{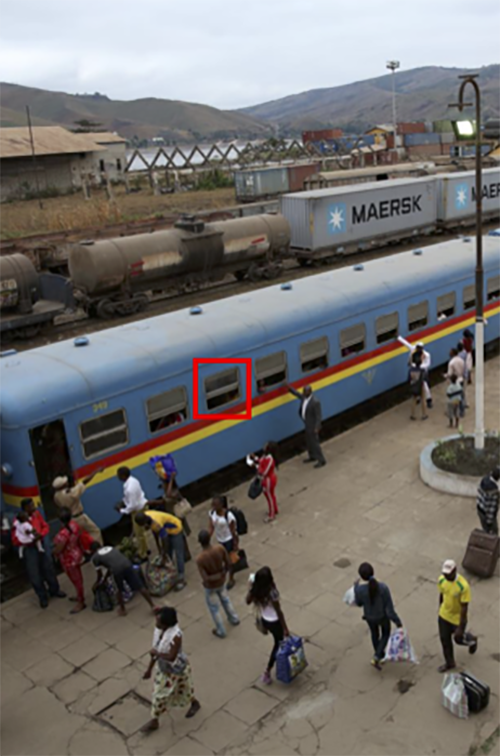}
\\
DIV2K 0850, degradation \ding{172}
\end{tabular}
\end{adjustbox}
\hspace{-0.46cm}
\begin{adjustbox}{valign=t}
\begin{tabular}{cccccc}
\includegraphics[width=0.18\textwidth, viewport=0 5 96 85, clip]{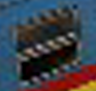} \hspace{-4mm} &
\includegraphics[width=0.18\textwidth, viewport=0 5 96 85, clip]{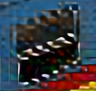} \hspace{-4mm} &
\includegraphics[width=0.18\textwidth, viewport=0 5 96 85, clip]{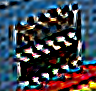} \hspace{-4mm} &
\includegraphics[width=0.18\textwidth, viewport=0 5 96 85, clip]{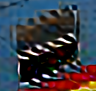} \hspace{-4mm} &
\includegraphics[width=0.18\textwidth, viewport=0 5 96 85, clip]{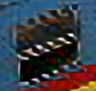} \hspace{-4mm} &
\includegraphics[width=0.18\textwidth, viewport=0 5 96 85, clip]{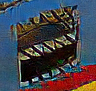} \hspace{-4mm} 
\\
Bicubic  \hspace{-4mm} &
RCAN  \hspace{-4mm} &
KernelGAN  \hspace{-4mm} &
IKC  \hspace{-4mm} &
DASR  \hspace{-4mm} &
FSSR  \hspace{-4mm}
\\
\includegraphics[width=0.18\textwidth, viewport=0 5 96 85, clip]{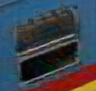} \hspace{-4mm} &
\includegraphics[width=0.18\textwidth, viewport=0 5 96 85, clip]{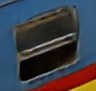} \hspace{-4mm} &
\includegraphics[width=0.18\textwidth, viewport=0 5 96 85, clip]{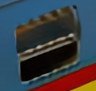} \hspace{-4mm} &
\includegraphics[width=0.18\textwidth, viewport=0 5 96 85, clip]{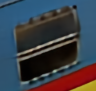} \hspace{-4mm} &
\includegraphics[width=0.18\textwidth, viewport=0 5 96 85, clip]{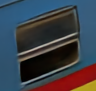} \hspace{-4mm} &
\includegraphics[width=0.18\textwidth, viewport=0 5 96 85, clip]{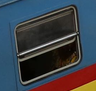} \hspace{-4mm}  
\\ 
USR-DA   \hspace{-4mm} &
PDM-SR  \hspace{-4mm} &
BSRNet  \hspace{-4mm} &
Real-ESRNet  \hspace{-4mm} &
Ours \hspace{-4mm} &
GT \hspace{-4mm}
\\
\end{tabular}
\end{adjustbox}
\\

% % % % one row

\hspace{-0.45cm}
\begin{adjustbox}{valign=t}
\begin{tabular}{c}
\includegraphics[width=0.211\textwidth, viewport=50  430 420 764, clip]{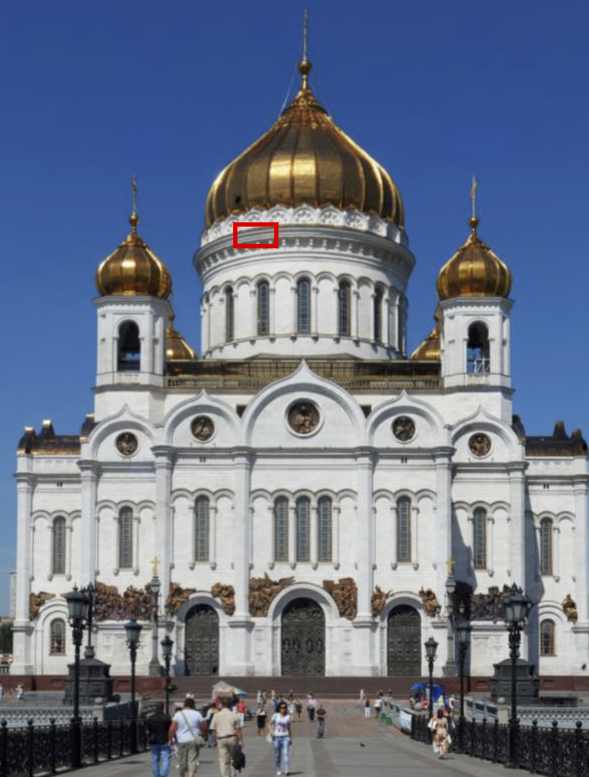}
\\
DIV2K 0879, degradation \ding{172}
\end{tabular}
\end{adjustbox}
\hspace{-0.46cm}
\begin{adjustbox}{valign=t}
\begin{tabular}{cccccc}
\includegraphics[width=0.18\textwidth]{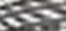} \hspace{-4mm} &
\includegraphics[width=0.18\textwidth]{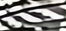} \hspace{-4mm} &
\includegraphics[width=0.18\textwidth]{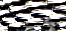} \hspace{-4mm} &
\includegraphics[width=0.18\textwidth]{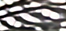} \hspace{-4mm} &
\includegraphics[width=0.18\textwidth]{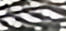} \hspace{-4mm} &
\includegraphics[width=0.18\textwidth]{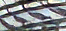} \hspace{-4mm} 
\\
Bicubic  \hspace{-4mm} &
RCAN  \hspace{-4mm} &
KernelGAN  \hspace{-4mm} &
IKC  \hspace{-4mm} &
DASR  \hspace{-4mm} &
FSSR  \hspace{-4mm}
\\
\includegraphics[width=0.18\textwidth]{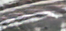} \hspace{-4mm} &
\includegraphics[width=0.18\textwidth]{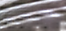} \hspace{-4mm} &
\includegraphics[width=0.18\textwidth]{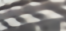} \hspace{-4mm} &
\includegraphics[width=0.18\textwidth]{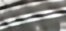} \hspace{-4mm} &
\includegraphics[width=0.18\textwidth]{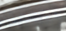} \hspace{-4mm} &
\includegraphics[width=0.18\textwidth]{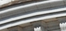} \hspace{-4mm}  
\\ 
USR-DA   \hspace{-4mm} &
PDM-SR  \hspace{-4mm} &
BSRNet  \hspace{-4mm} &
Real-ESRNet  \hspace{-4mm} &
Ours \hspace{-4mm} &
GT \hspace{-4mm}
\\
\end{tabular}
\end{adjustbox}
\\

% % % % one row

\hspace{-0.45cm}
\begin{adjustbox}{valign=t}
\begin{tabular}{c}
\includegraphics[width=0.211\textwidth, viewport=20  20 360 425, clip]{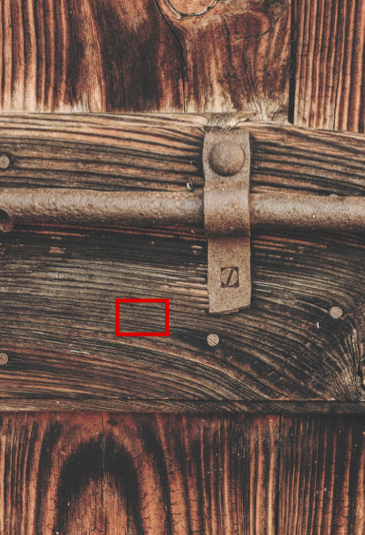}
\\
DIV2K 0885, degradation \ding{172}
\end{tabular}
\end{adjustbox}
\hspace{-0.46cm}
\begin{adjustbox}{valign=t}
\begin{tabular}{cccccc}
\includegraphics[width=0.18\textwidth]{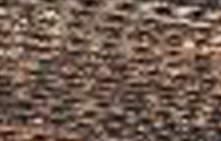} \hspace{-4mm} &
\includegraphics[width=0.18\textwidth]{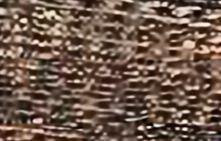} \hspace{-4mm} &
\includegraphics[width=0.18\textwidth]{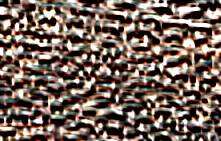} \hspace{-4mm} &
\includegraphics[width=0.18\textwidth]{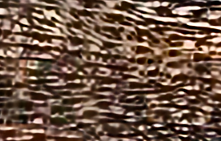} \hspace{-4mm} &
\includegraphics[width=0.18\textwidth]{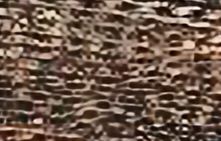} \hspace{-4mm} &
\includegraphics[width=0.18\textwidth]{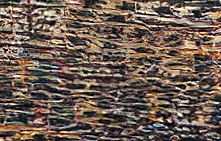} \hspace{-4mm} 
\\
Bicubic  \hspace{-4mm} &
RCAN  \hspace{-4mm} &
KernelGAN  \hspace{-4mm} &
IKC  \hspace{-4mm} &
DASR  \hspace{-4mm} &
FSSR  \hspace{-4mm}
\\
\includegraphics[width=0.18\textwidth]{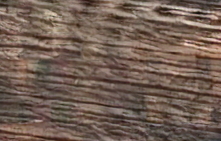} \hspace{-4mm} &
\includegraphics[width=0.18\textwidth]{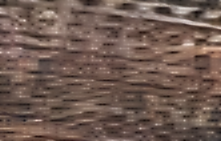} \hspace{-4mm} &
\includegraphics[width=0.18\textwidth]{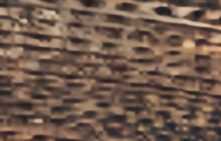} \hspace{-4mm} &
\includegraphics[width=0.18\textwidth]{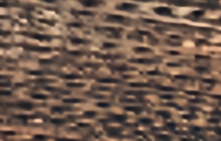} \hspace{-4mm} &
\includegraphics[width=0.18\textwidth]{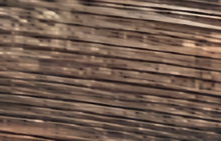} \hspace{-4mm} &
\includegraphics[width=0.18\textwidth]{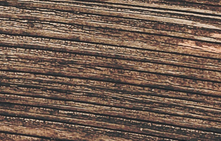} \hspace{-4mm}  
\\ 
USR-DA   \hspace{-4mm} &
PDM-SR  \hspace{-4mm} &
BSRNet  \hspace{-4mm} &
Real-ESRNet  \hspace{-4mm} &
Ours \hspace{-4mm} &
GT \hspace{-4mm}
\\
\end{tabular}
\end{adjustbox}
\\

% % % % one row

% % one row

\hspace{-0.45cm}
\begin{adjustbox}{valign=t}
\begin{tabular}{c}
\includegraphics[width=0.211\textwidth, viewport=20  220 175 431, clip]{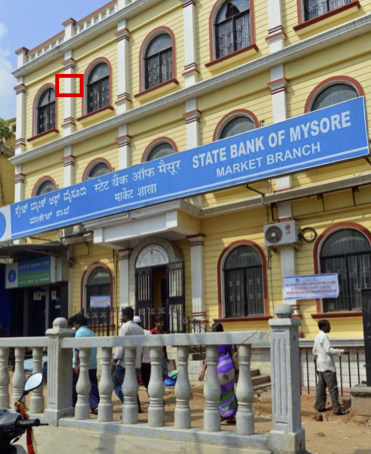}
\\
DIV2K 0891, degradation \ding{172}
\end{tabular}
\end{adjustbox}
\hspace{-0.46cm}
\begin{adjustbox}{valign=t}
\begin{tabular}{cccccc}
\includegraphics[width=0.18\textwidth]{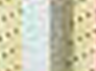} \hspace{-4mm} &
\includegraphics[width=0.18\textwidth]{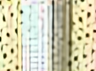} \hspace{-4mm} &
\includegraphics[width=0.18\textwidth]{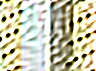} \hspace{-4mm} &
\includegraphics[width=0.18\textwidth]{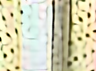} \hspace{-4mm} &
\includegraphics[width=0.18\textwidth]{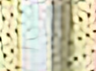} \hspace{-4mm} &
\includegraphics[width=0.18\textwidth]{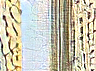} \hspace{-4mm} 
\\
Bicubic  \hspace{-4mm} &
RCAN  \hspace{-4mm} &
KernelGAN  \hspace{-4mm} &
IKC  \hspace{-4mm} &
DASR  \hspace{-4mm} &
FSSR  \hspace{-4mm}
\\
\includegraphics[width=0.18\textwidth]{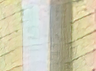} \hspace{-4mm} &
\includegraphics[width=0.18\textwidth]{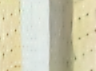} \hspace{-4mm} &
\includegraphics[width=0.18\textwidth]{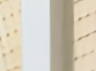} \hspace{-4mm} &
\includegraphics[width=0.18\textwidth]{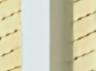} \hspace{-4mm} &
\includegraphics[width=0.18\textwidth]{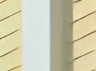} \hspace{-4mm} &
\includegraphics[width=0.18\textwidth]{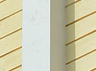} \hspace{-4mm}  
\\ 
USR-DA   \hspace{-4mm} &
PDM-SR  \hspace{-4mm} &
BSRNet  \hspace{-4mm} &
Real-ESRNet  \hspace{-4mm} &
Ours \hspace{-4mm} &
GT \hspace{-4mm}
\\
\end{tabular}
\end{adjustbox}
\\

\end{tabular}}
\vspace{-3mm}
\caption{SR Results of synthesized testing images with scale factor $\times$4.}
\label{fig:synthesizedsupp1}
\vspace{-3mm}
\end{figure*}

\begin{figure*}[t]
%\newlength-4mm
%\setlength{-4mm}{-0.4cm}
\scriptsize
\centering
\resizebox{1\textwidth}{!}{
\begin{tabular}{ccc}

% % % % one row

\hspace{-0.45cm}
\begin{adjustbox}{valign=t}
\begin{tabular}{c}
\includegraphics[width=0.211\textwidth, viewport=0  100 361 507, clip]{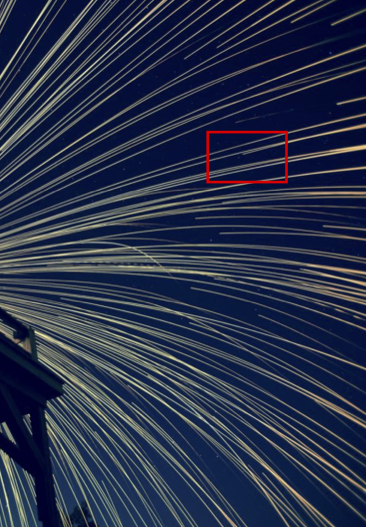}
\\
DIV2K 0828, degradation \ding{173}
\end{tabular}
\end{adjustbox}
\hspace{-0.46cm}
\begin{adjustbox}{valign=t}
\begin{tabular}{cccccc}
\includegraphics[width=0.18\textwidth]{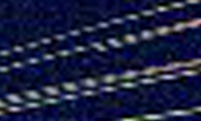} \hspace{-4mm} &
\includegraphics[width=0.18\textwidth]{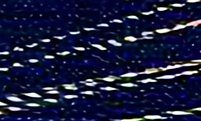} \hspace{-4mm} &
\includegraphics[width=0.18\textwidth]{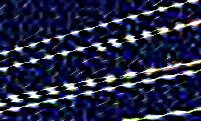} \hspace{-4mm} &
\includegraphics[width=0.18\textwidth]{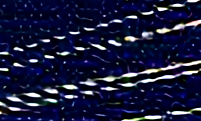} \hspace{-4mm} &
\includegraphics[width=0.18\textwidth]{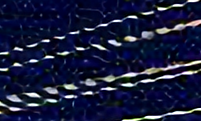} \hspace{-4mm} &
\includegraphics[width=0.18\textwidth]{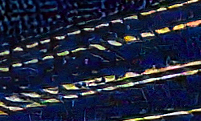} \hspace{-4mm} 
\\
Bicubic  \hspace{-4mm} &
RCAN  \hspace{-4mm} &
KernelGAN  \hspace{-4mm} &
IKC  \hspace{-4mm} &
DASR  \hspace{-4mm} &
FSSR  \hspace{-4mm}
\\
\includegraphics[width=0.18\textwidth]{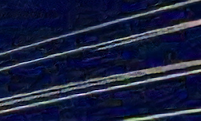} \hspace{-4mm} &
\includegraphics[width=0.18\textwidth]{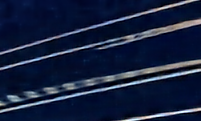} \hspace{-4mm} &
\includegraphics[width=0.18\textwidth]{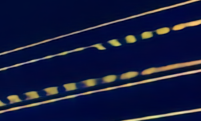} \hspace{-4mm} &
\includegraphics[width=0.18\textwidth]{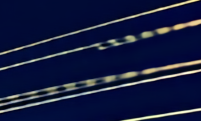} \hspace{-4mm} &
\includegraphics[width=0.18\textwidth]{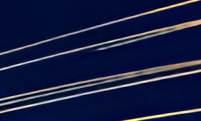} \hspace{-4mm} &
\includegraphics[width=0.18\textwidth]{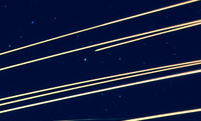} \hspace{-4mm}  
\\ 
USR-DA   \hspace{-4mm} &
PDM-SR  \hspace{-4mm} &
BSRNet  \hspace{-4mm} &
Real-ESRNet  \hspace{-4mm} &
Ours \hspace{-4mm} &
GT \hspace{-4mm}
\\
\end{tabular}
\end{adjustbox}
\\

% % % % one row

% % % % one row

\hspace{-0.45cm}
\begin{adjustbox}{valign=t}
\begin{tabular}{c}
\includegraphics[width=0.211\textwidth, viewport=0  0 380 473, clip]{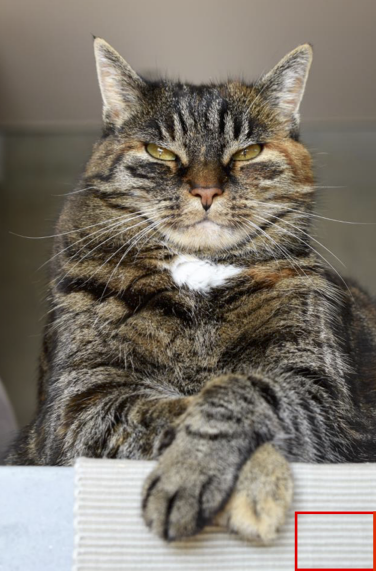}
\\
DIV2K 0869, degradation \ding{174}
\end{tabular}
\end{adjustbox}
\hspace{-0.46cm}
\begin{adjustbox}{valign=t}
\begin{tabular}{cccccc}
\includegraphics[width=0.18\textwidth]{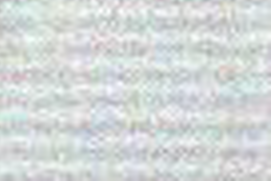} \hspace{-4mm} &
\includegraphics[width=0.18\textwidth]{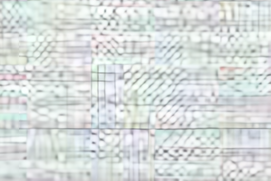} \hspace{-4mm} &
\includegraphics[width=0.18\textwidth]{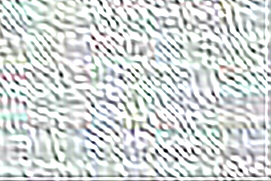} \hspace{-4mm} &
\includegraphics[width=0.18\textwidth]{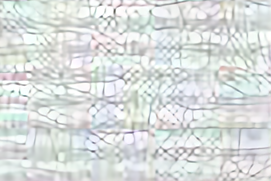} \hspace{-4mm} &
\includegraphics[width=0.18\textwidth]{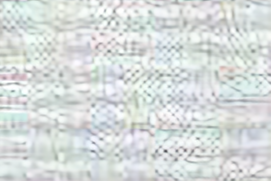} \hspace{-4mm} &
\includegraphics[width=0.18\textwidth]{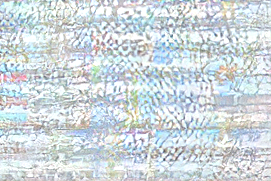} \hspace{-4mm} 
\\
Bicubic  \hspace{-4mm} &
RCAN  \hspace{-4mm} &
KernelGAN  \hspace{-4mm} &
IKC  \hspace{-4mm} &
DASR  \hspace{-4mm} &
FSSR  \hspace{-4mm}
\\
\includegraphics[width=0.18\textwidth]{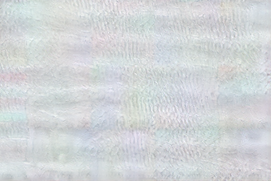} \hspace{-4mm} &
\includegraphics[width=0.18\textwidth]{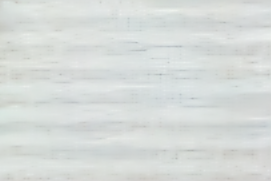} \hspace{-4mm} &
\includegraphics[width=0.18\textwidth]{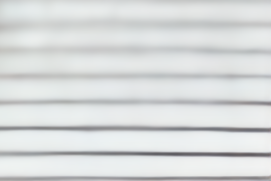} \hspace{-4mm} &
\includegraphics[width=0.18\textwidth]{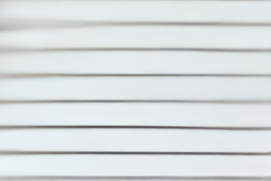} \hspace{-4mm} &
\includegraphics[width=0.18\textwidth]{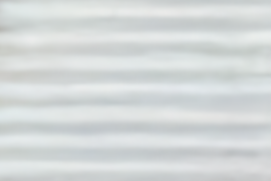} \hspace{-4mm} &
\includegraphics[width=0.18\textwidth]{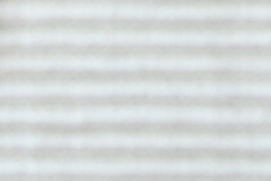} \hspace{-4mm}  
\\ 
USR-DA   \hspace{-4mm} &
PDM-SR  \hspace{-4mm} &
BSRNet  \hspace{-4mm} &
Real-ESRNet  \hspace{-4mm} &
Ours \hspace{-4mm} &
GT \hspace{-4mm}
\\
\end{tabular}
\end{adjustbox}
\\

% % % % one row

\end{tabular}}
\vspace{-3mm}
\caption{SR Results of synthesized testing images with scale factor $\times$4.}
\label{fig:synthesizedsupp2}
\vspace{-3mm}
\end{figure*}

\begin{figure*}[t]
%\newlength-4mm
%\setlength{-4mm}{-0.4cm}
\scriptsize
\centering
\resizebox{1\textwidth}{!}{
\begin{tabular}{ccc}

% % one row

\hspace{-0.45cm}
\begin{adjustbox}{valign=t}
\begin{tabular}{c}
\includegraphics[width=0.211\textwidth, viewport=0  0 300 286, clip]{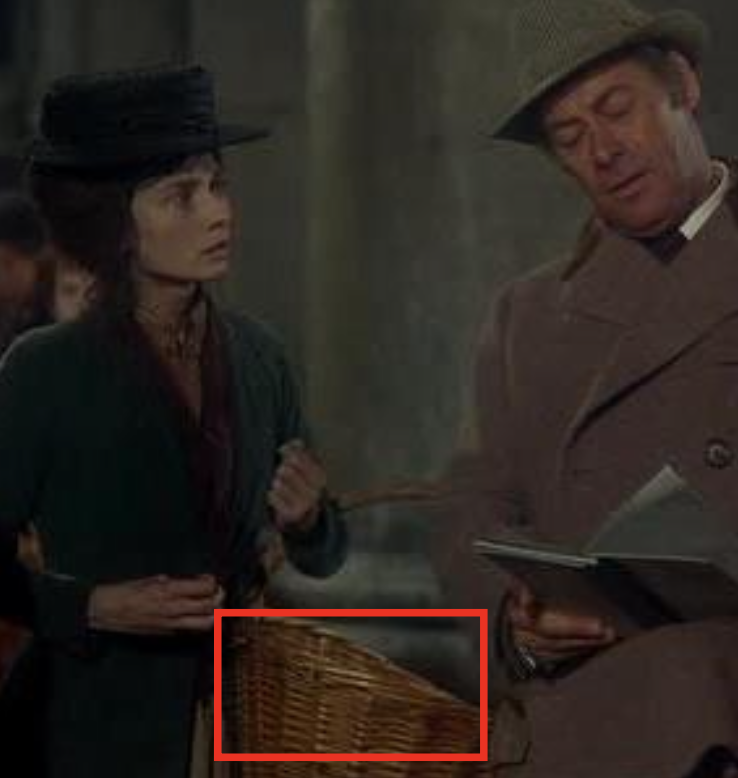}
\\
Lady
\end{tabular}
\end{adjustbox}
\hspace{-0.46cm}
\begin{adjustbox}{valign=t}
\begin{tabular}{cccccc}
\includegraphics[width=0.18\textwidth]{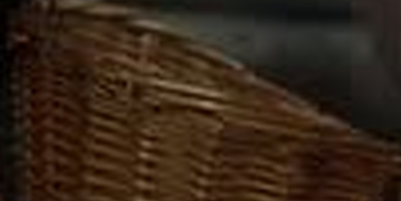} \hspace{-4mm} &
\includegraphics[width=0.18\textwidth]{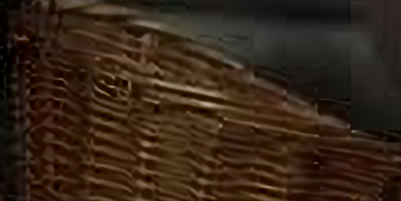} \hspace{-4mm} &
\includegraphics[width=0.18\textwidth]{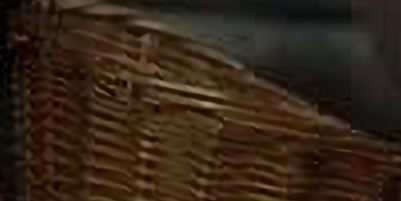} \hspace{-4mm} &
\includegraphics[width=0.18\textwidth]{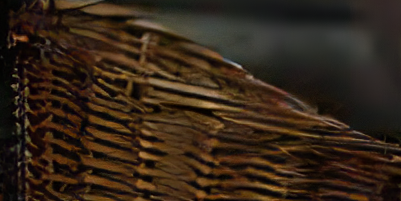} \hspace{-4mm} 
\\
Bicubic  \hspace{-4mm} &
IKC  \hspace{-4mm} &
DASR  \hspace{-4mm} &
FSSR  \hspace{-4mm}
\\
\includegraphics[width=0.18\textwidth]{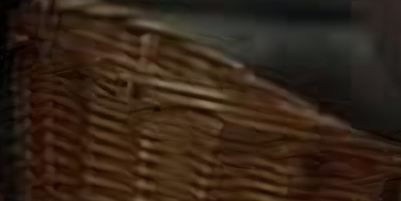} \hspace{-4mm} &
\includegraphics[width=0.18\textwidth]{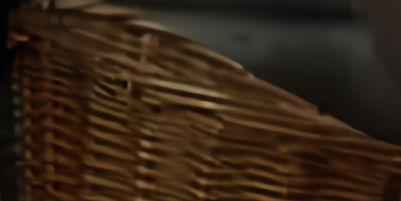} \hspace{-4mm} &
\includegraphics[width=0.18\textwidth]{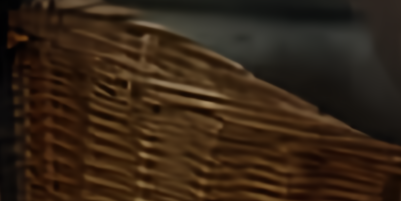} \hspace{-4mm} &
\includegraphics[width=0.18\textwidth]{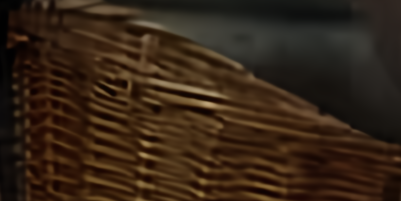} \hspace{-4mm}  
\\ 
USR-DA   \hspace{-4mm} &
BSRNet  \hspace{-4mm} &
Real-ESRNet \hspace{-4mm} &
Ours \hspace{-4mm}
\\
\end{tabular}
\end{adjustbox}
\\

% % % % one row

\hspace{-0.45cm}
\begin{adjustbox}{valign=t}
\begin{tabular}{c}
\includegraphics[width=0.211\textwidth, viewport=40  50 200 275, clip]{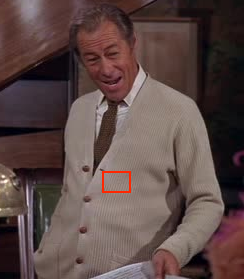}
\\
Lady
\end{tabular}
\end{adjustbox}
\hspace{-0.46cm}
\begin{adjustbox}{valign=t}
\begin{tabular}{cccccc}
\includegraphics[width=0.18\textwidth, viewport=0  50 171 181, clip]{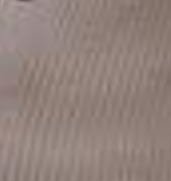} \hspace{-4mm} &
\includegraphics[width=0.18\textwidth, viewport=0  50 171 181, clip]{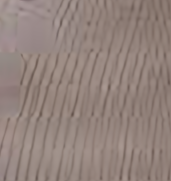} \hspace{-4mm} &
\includegraphics[width=0.18\textwidth, viewport=0  50 171 181, clip]{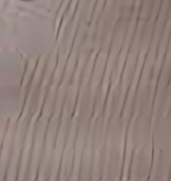} \hspace{-4mm} &
\includegraphics[width=0.18\textwidth, viewport=0  50 171 181, clip]{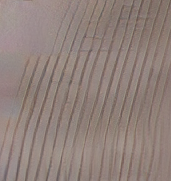} \hspace{-4mm} 
\\
Bicubic  \hspace{-4mm} &
IKC  \hspace{-4mm} &
DASR  \hspace{-4mm} &
FSSR  \hspace{-4mm}
\\
\includegraphics[width=0.18\textwidth, viewport=0  50 171 181, clip]{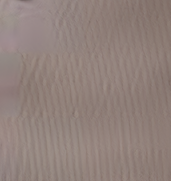} \hspace{-4mm} &
\includegraphics[width=0.18\textwidth, viewport=0  50 171 181, clip]{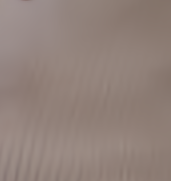} \hspace{-4mm} &
\includegraphics[width=0.18\textwidth, viewport=0  50 171 181, clip]{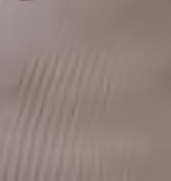} \hspace{-4mm} &
\includegraphics[width=0.18\textwidth, viewport=0  50 171 181, clip]{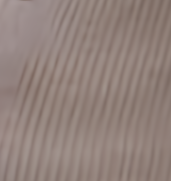} \hspace{-4mm}  
\\ 
USR-DA   \hspace{-4mm} &
BSRNet  \hspace{-4mm} &
Real-ESRNet \hspace{-4mm} &
Ours \hspace{-4mm}
\\
\end{tabular}
\end{adjustbox}
\\

% % % % one row

\hspace{-0.45cm}
\begin{adjustbox}{valign=t}
\begin{tabular}{c}
\includegraphics[width=0.211\textwidth, viewport=0  70 227 250, clip]{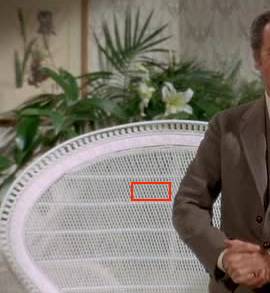}
\\

% Lady 0468
Lady
\end{tabular}
\end{adjustbox}
\hspace{-0.46cm}
\begin{adjustbox}{valign=t}
\begin{tabular}{cccccc}
\includegraphics[width=0.18\textwidth, viewport=0  3 131 56, clip]{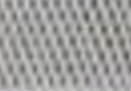} \hspace{-4mm} &
\includegraphics[width=0.18\textwidth, viewport=0  3 131 56, clip]{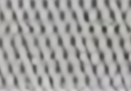} \hspace{-4mm} &
\includegraphics[width=0.18\textwidth, viewport=0  3 131 56, clip]{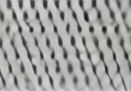} \hspace{-4mm} &
\includegraphics[width=0.18\textwidth, viewport=0  3 131 56, clip]{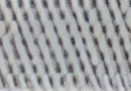} \hspace{-4mm} 
\\
Bicubic  \hspace{-4mm} &
IKC  \hspace{-4mm} &
DASR  \hspace{-4mm} &
FSSR  \hspace{-4mm}
\\
\includegraphics[width=0.18\textwidth, viewport=0  3 131 56, clip]{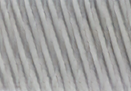} \hspace{-4mm} &
\includegraphics[width=0.18\textwidth, viewport=0  3 131 56, clip]{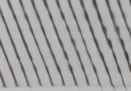} \hspace{-4mm} &
\includegraphics[width=0.18\textwidth, viewport=0  3 131 56, clip]{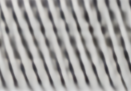} \hspace{-4mm} &
\includegraphics[width=0.18\textwidth, viewport=0  3 131 56, clip]{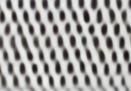} \hspace{-4mm}  
\\ 
USR-DA   \hspace{-4mm} &
BSRNet  \hspace{-4mm} &
Real-ESRNet \hspace{-4mm} &
Ours \hspace{-4mm}
\\
\end{tabular}
\end{adjustbox}
\\
% % % % one row

\hspace{-0.45cm}
\begin{adjustbox}{valign=t}
\begin{tabular}{c}
\includegraphics[width=0.211\textwidth, viewport=50  100 450 365, clip]{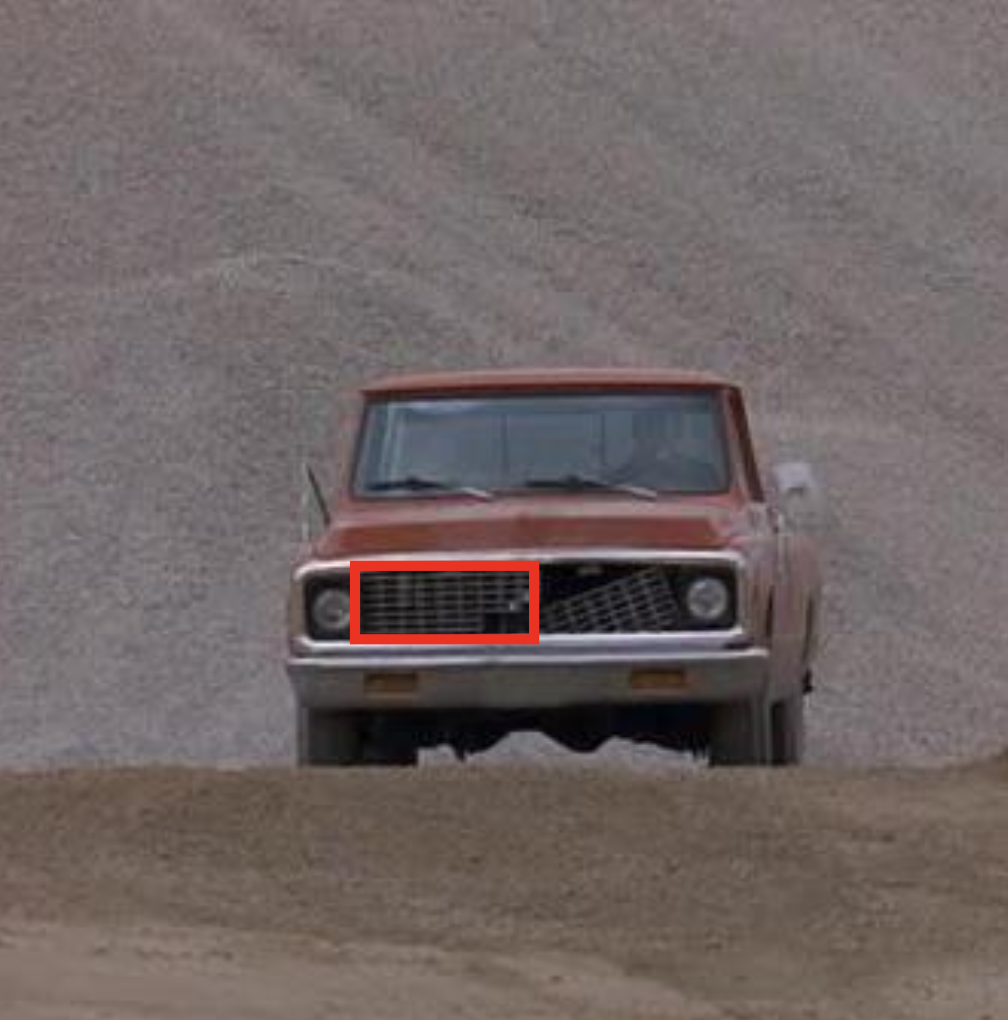}
\\
% Groundhog 0305
Groundhog
\end{tabular}
\end{adjustbox}
\hspace{-0.46cm}
\begin{adjustbox}{valign=t}
\begin{tabular}{cccccc}
\includegraphics[width=0.18\textwidth]{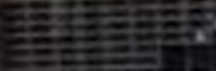} \hspace{-4mm} &
\includegraphics[width=0.18\textwidth]{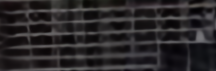} \hspace{-4mm} &
\includegraphics[width=0.18\textwidth]{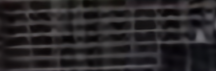} \hspace{-4mm} &
\includegraphics[width=0.18\textwidth]{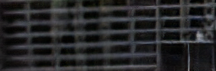} \hspace{-4mm} 
\\
Bicubic  \hspace{-4mm} &
IKC  \hspace{-4mm} &
DASR  \hspace{-4mm} &
FSSR  \hspace{-4mm}
\\
\includegraphics[width=0.18\textwidth]{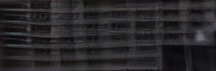} \hspace{-4mm} &
\includegraphics[width=0.18\textwidth]{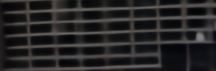} \hspace{-4mm} &
\includegraphics[width=0.18\textwidth]{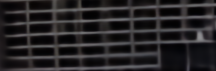} \hspace{-4mm} &
\includegraphics[width=0.18\textwidth]{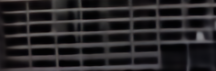} \hspace{-4mm}  
\\ 
USR-DA   \hspace{-4mm} &
BSRNet  \hspace{-4mm} &
Real-ESRNet \hspace{-4mm} &
Ours \hspace{-4mm}
\\
\end{tabular}
\end{adjustbox}
\\

% % % one row

\hspace{-0.45cm}
\begin{adjustbox}{valign=t}
\begin{tabular}{c}
\includegraphics[width=0.211\textwidth, viewport=170  0 370 263, clip]{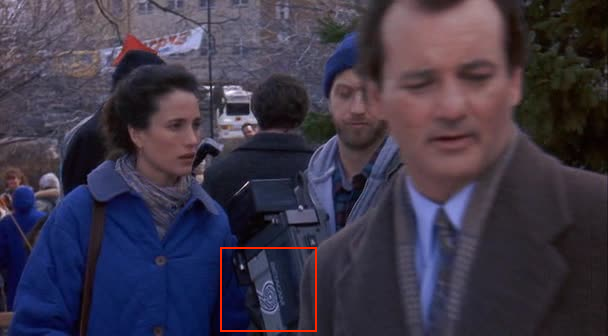}
\\
Groundhog
\end{tabular}
\end{adjustbox}
\hspace{-0.46cm}
\begin{adjustbox}{valign=t}
\begin{tabular}{cccccc}
\includegraphics[width=0.18\textwidth, viewport=0  0 450 320, clip]{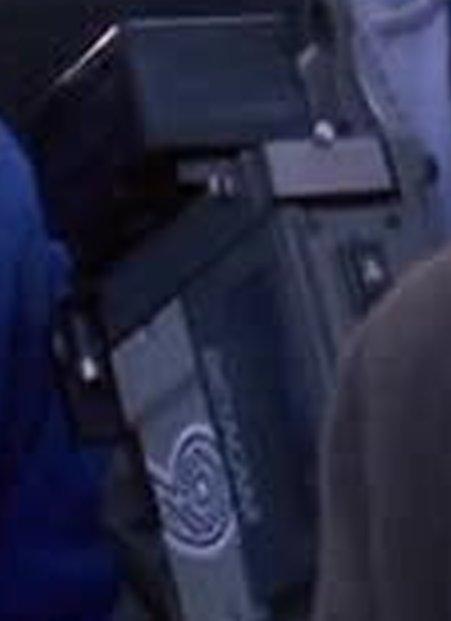} \hspace{-4mm} &
\includegraphics[width=0.18\textwidth, viewport=0  0 450 320, clip]{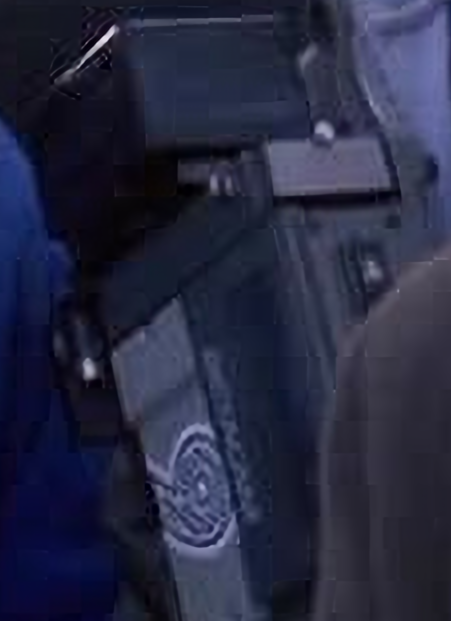} \hspace{-4mm} &
\includegraphics[width=0.18\textwidth, viewport=0  0 450 320, clip]{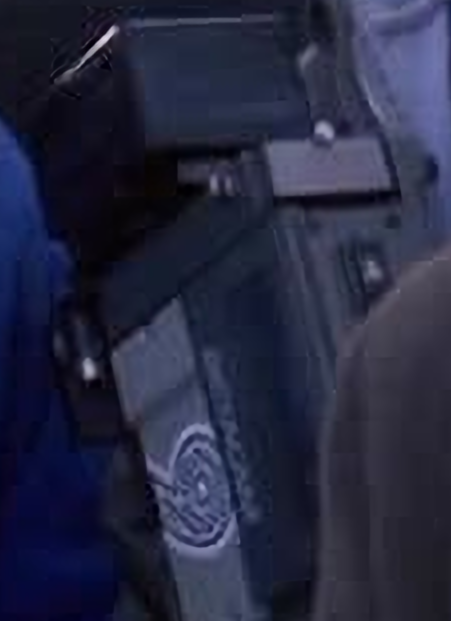} \hspace{-4mm} &
\includegraphics[width=0.18\textwidth, viewport=0  0 450 320, clip]{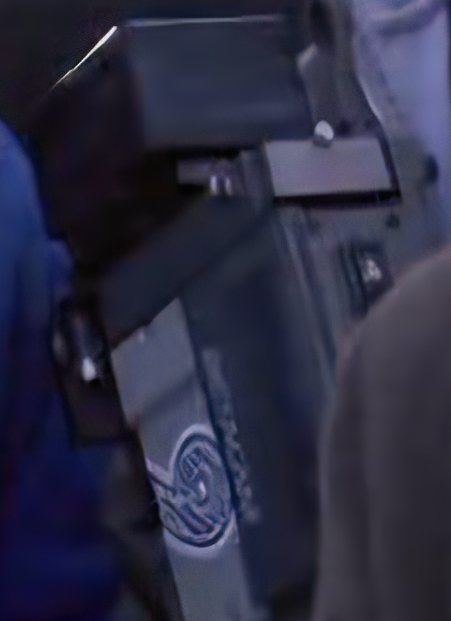} \hspace{-4mm} 
\\
Bicubic  \hspace{-4mm} &
IKC  \hspace{-4mm} &
DASR  \hspace{-4mm} &
FSSR  \hspace{-4mm}
\\
\includegraphics[width=0.18\textwidth, viewport=0  0 450 320, clip]{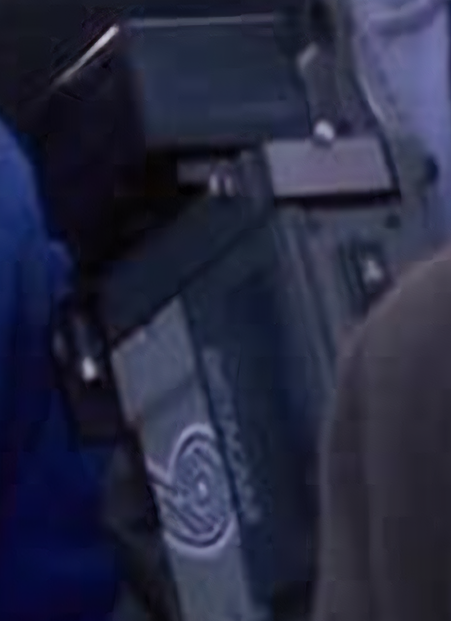} \hspace{-4mm} &
\includegraphics[width=0.18\textwidth, viewport=0  0 450 320, clip]{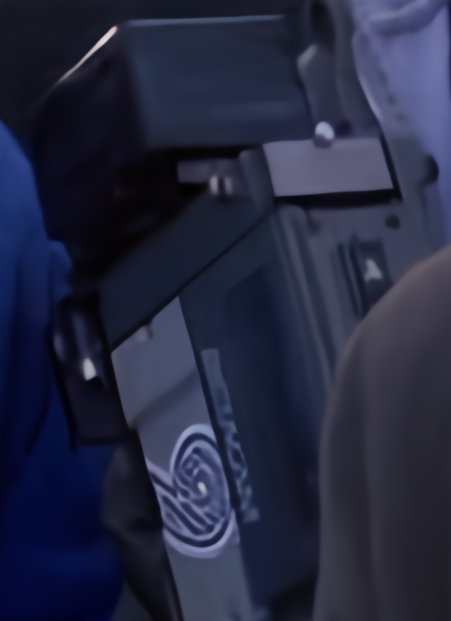} \hspace{-4mm} &
\includegraphics[width=0.18\textwidth, viewport=0  0 450 320, clip]{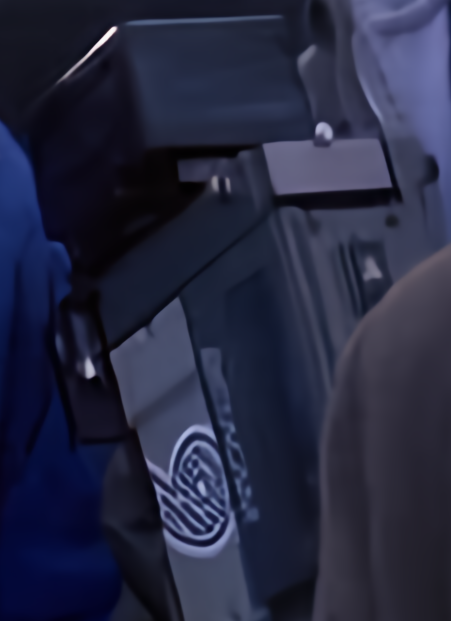} \hspace{-4mm} &
\includegraphics[width=0.18\textwidth, viewport=0  0 450 320, clip]{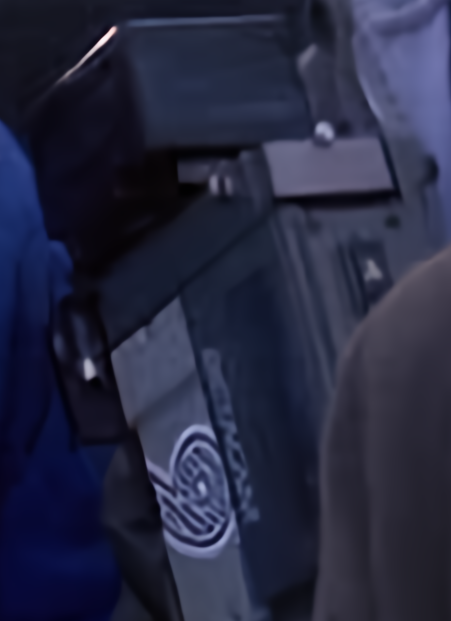} \hspace{-4mm}  
\\ 
USR-DA   \hspace{-4mm} &
BSRNet  \hspace{-4mm} &
Real-ESRNet \hspace{-4mm} &
Ours \hspace{-4mm}
\\
\end{tabular}
\end{adjustbox}
\\

\end{tabular}}
\vspace{-3mm}
\caption{SR results of real-world images with scale factor $\times$4.}
\label{fig:realsupp1}
\vspace{-3mm}
\end{figure*}

%%%%%%%%%%%%%%%%%%%%%%%%%%%%%%%%%%%%%%%%%%%%%%%%%%%%%%%%%%%%%%%%%%%%%%%%%%%%%%%
%%%%%%%%%%%%%%%%%%%%%%%%%%%%%%%%%%%%%%%%%%%%%%%%%%%%%%%%%%%%%%%%%%%%%%%%%%%%%%%

\begin{figure*}[t]
%\newlength-4mm
%\setlength{-4mm}{-0.4cm}
\scriptsize
\centering
\resizebox{1\textwidth}{!}{
\begin{tabular}{ccc}

% % % % % one row

\hspace{-0.45cm}
\begin{adjustbox}{valign=t}
\begin{tabular}{c}
\includegraphics[width=0.211\textwidth, viewport=166  100 500 500, clip]{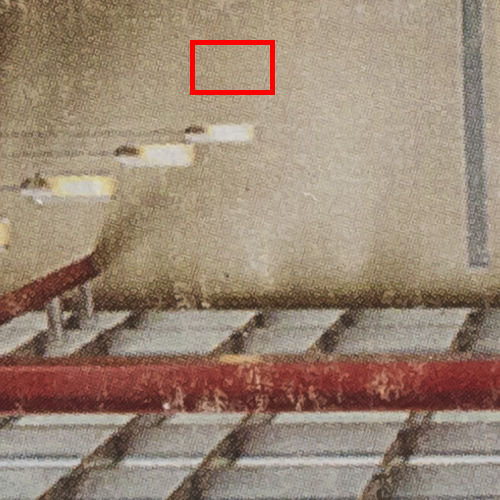}
\\
Nikon
\end{tabular}
\end{adjustbox}
\hspace{-0.46cm}
\begin{adjustbox}{valign=t}
\begin{tabular}{cccccc}
\includegraphics[width=0.18\textwidth]{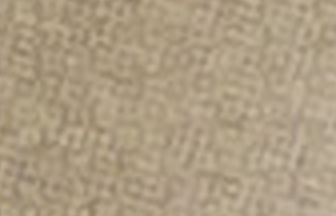} \hspace{-4mm} &
\includegraphics[width=0.18\textwidth]{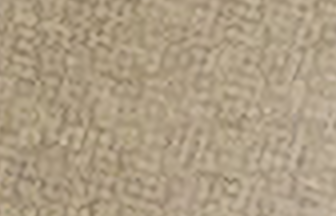} \hspace{-4mm} &
\includegraphics[width=0.18\textwidth]{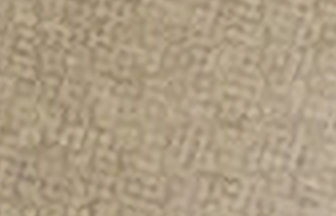} \hspace{-4mm} &
\includegraphics[width=0.18\textwidth]{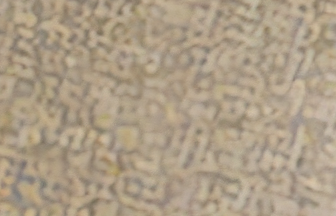} \hspace{-4mm} 
\\
Bicubic  \hspace{-4mm} &
IKC  \hspace{-4mm} &
DASR  \hspace{-4mm} &
FSSR  \hspace{-4mm}
\\
\includegraphics[width=0.18\textwidth]{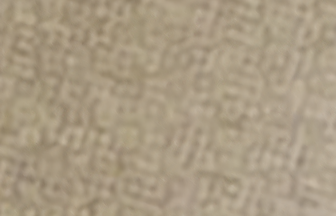} \hspace{-4mm} &
\includegraphics[width=0.18\textwidth]{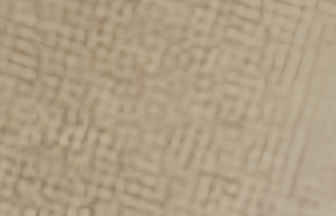} \hspace{-4mm} &
\includegraphics[width=0.18\textwidth]{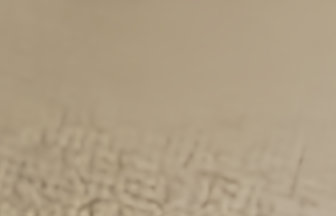} \hspace{-4mm} &
\includegraphics[width=0.18\textwidth]{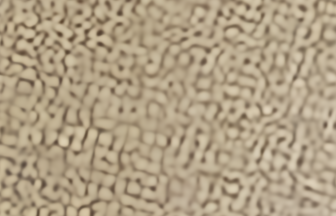} \hspace{-4mm}  
\\ 
USR-DA   \hspace{-4mm} &
BSRNet  \hspace{-4mm} &
Real-ESRNet \hspace{-4mm} &
Ours \hspace{-4mm}
\\
\end{tabular}
\end{adjustbox}
\\

% % % % one row

% % one row

\hspace{-0.45cm}
\begin{adjustbox}{valign=t}
\begin{tabular}{c}
\includegraphics[width=0.211\textwidth, viewport=52  61 500 500, clip]{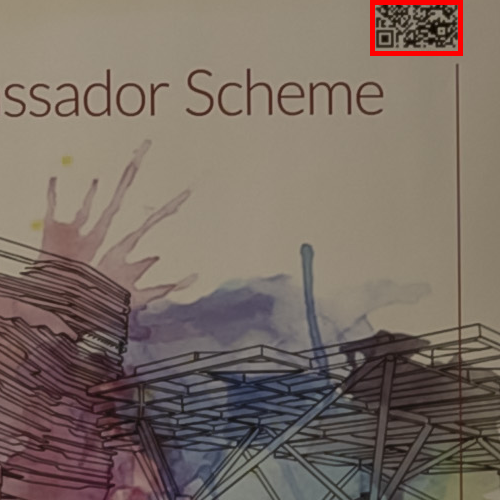}
\\
Nikon
\end{tabular}
\end{adjustbox}
\hspace{-0.46cm}
\begin{adjustbox}{valign=t}
\begin{tabular}{cccccc}
\includegraphics[width=0.18\textwidth, viewport=0  29 371 220, clip]{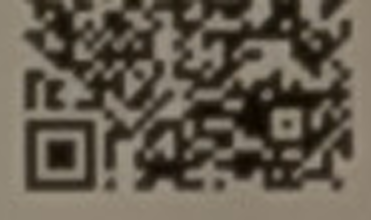} \hspace{-4mm} &
\includegraphics[width=0.18\textwidth, viewport=0  29 371 220, clip]{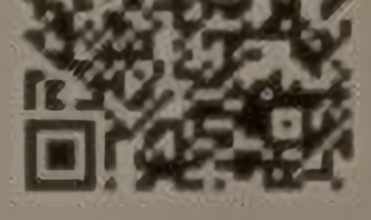} \hspace{-4mm} &
\includegraphics[width=0.18\textwidth, viewport=0  29 371 220, clip]{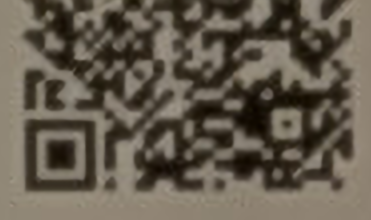} \hspace{-4mm} &
\includegraphics[width=0.18\textwidth, viewport=0  29 371 220, clip]{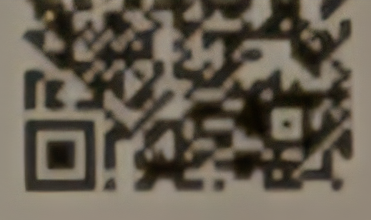} \hspace{-4mm} 
\\
Bicubic  \hspace{-4mm} &
IKC  \hspace{-4mm} &
DASR  \hspace{-4mm} &
FSSR  \hspace{-4mm}
\\
\includegraphics[width=0.18\textwidth, viewport=0  29 371 220, clip]{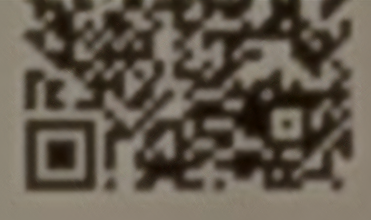} \hspace{-4mm} &
\includegraphics[width=0.18\textwidth, viewport=0  29 371 220, clip]{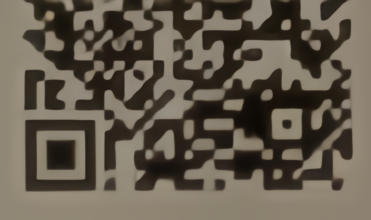} \hspace{-4mm} &
\includegraphics[width=0.18\textwidth, viewport=0  29 371 220, clip]{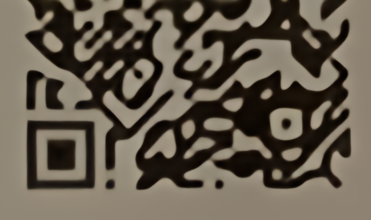} \hspace{-4mm} &
\includegraphics[width=0.18\textwidth, viewport=0  29 371 220, clip]{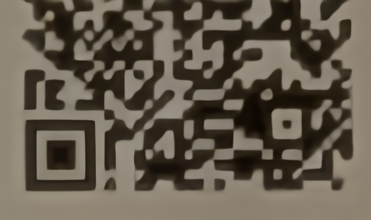} \hspace{-4mm}  
\\ 
USR-DA   \hspace{-4mm} &
BSRNet  \hspace{-4mm} &
Real-ESRNet \hspace{-4mm} &
Ours \hspace{-4mm}
\\
\end{tabular}
\end{adjustbox}
\\

% % % % one row

\hspace{-0.45cm}
\begin{adjustbox}{valign=t}
\begin{tabular}{c}
\includegraphics[width=0.211\textwidth, viewport=120  1 500 500, clip]{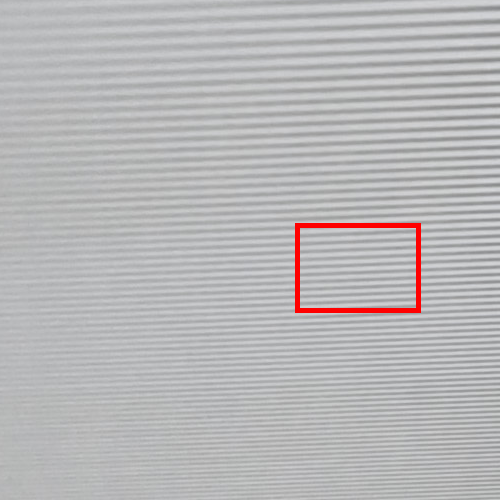}
\\
Nikon
\end{tabular}
\end{adjustbox}
\hspace{-0.46cm}
\begin{adjustbox}{valign=t}
\begin{tabular}{cccccc}
\includegraphics[width=0.18\textwidth]{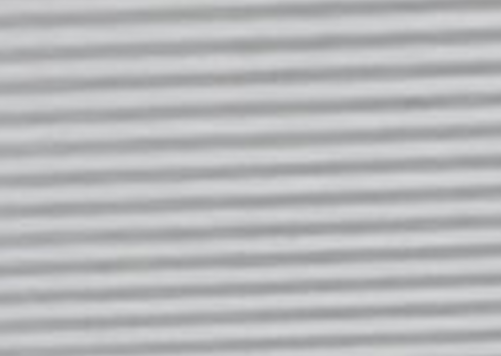} \hspace{-4mm} &
\includegraphics[width=0.18\textwidth]{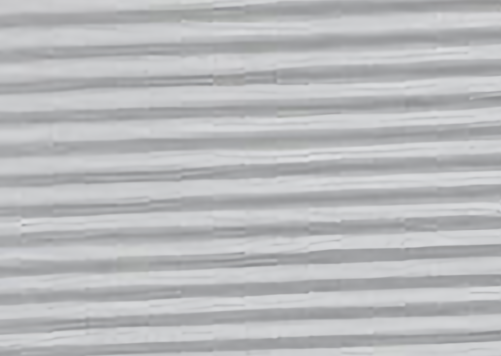} \hspace{-4mm} &
\includegraphics[width=0.18\textwidth]{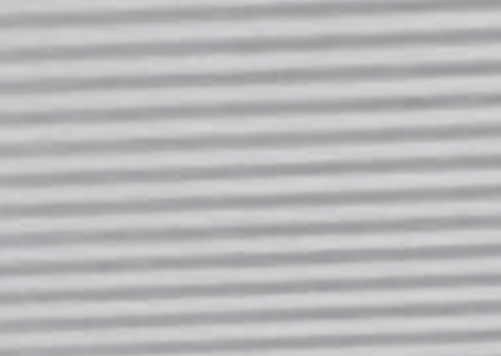} \hspace{-4mm} &
\includegraphics[width=0.18\textwidth]{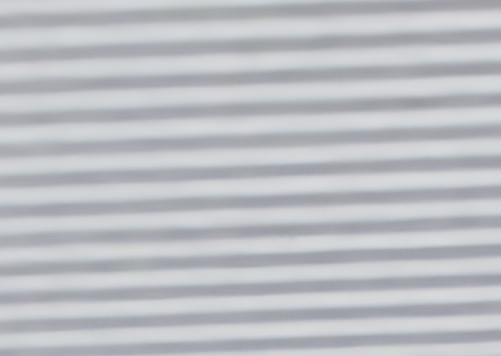} \hspace{-4mm} 
\\
Bicubic  \hspace{-4mm} &
IKC  \hspace{-4mm} &
DASR  \hspace{-4mm} &
FSSR  \hspace{-4mm}
\\
\includegraphics[width=0.18\textwidth]{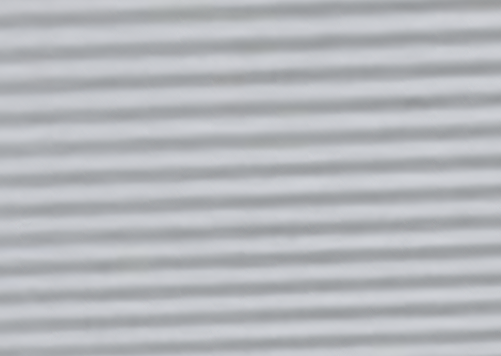} \hspace{-4mm} &
\includegraphics[width=0.18\textwidth]{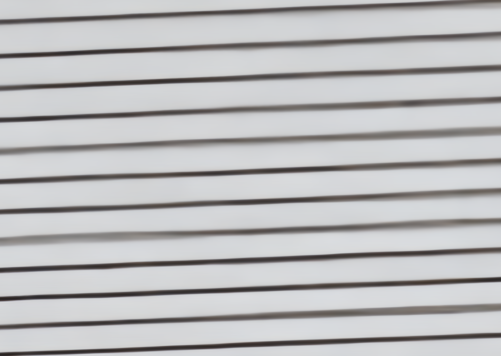} \hspace{-4mm} &
\includegraphics[width=0.18\textwidth]{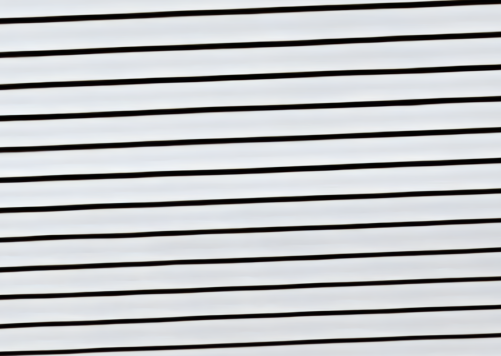} \hspace{-4mm} &
\includegraphics[width=0.18\textwidth]{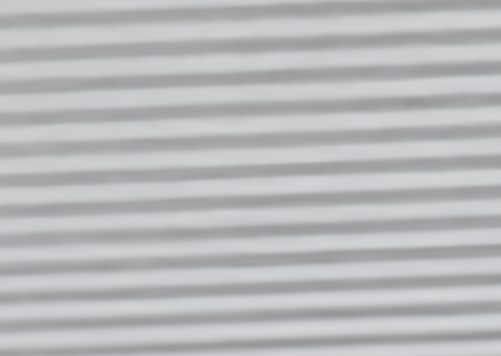} \hspace{-4mm}  
\\ 
USR-DA   \hspace{-4mm} &
BSRNet  \hspace{-4mm} &
Real-ESRNet \hspace{-4mm} &
Ours \hspace{-4mm}
\\
\end{tabular}
\end{adjustbox}
\\

% % % % one row

\hspace{-0.45cm}
\begin{adjustbox}{valign=t}
\begin{tabular}{c}
\includegraphics[width=0.211\textwidth, viewport=0  0 340 494, clip]{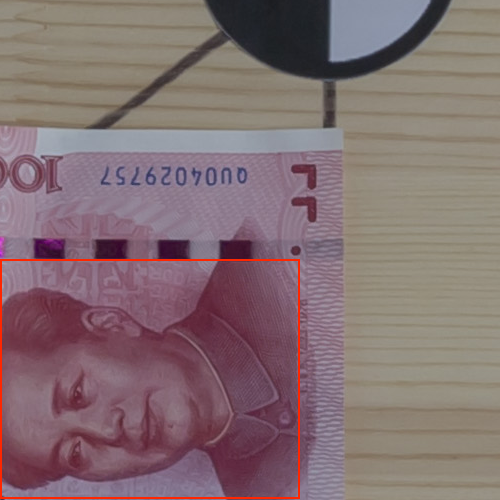}
\\
Canon
\end{tabular}
\end{adjustbox}
\hspace{-0.46cm}
\begin{adjustbox}{valign=t}
\begin{tabular}{cccccc}
\includegraphics[width=0.18\textwidth]{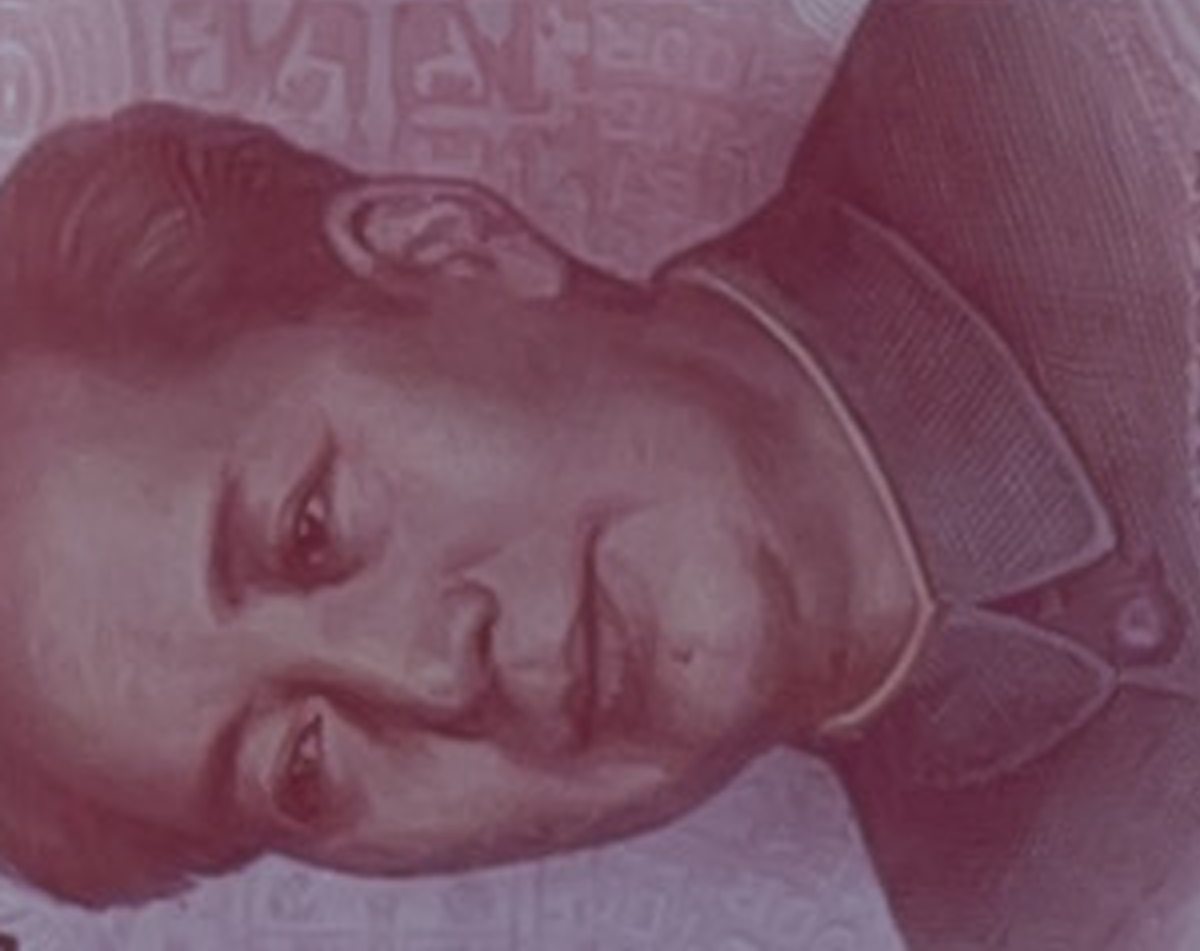} \hspace{-4mm} &
\includegraphics[width=0.18\textwidth]{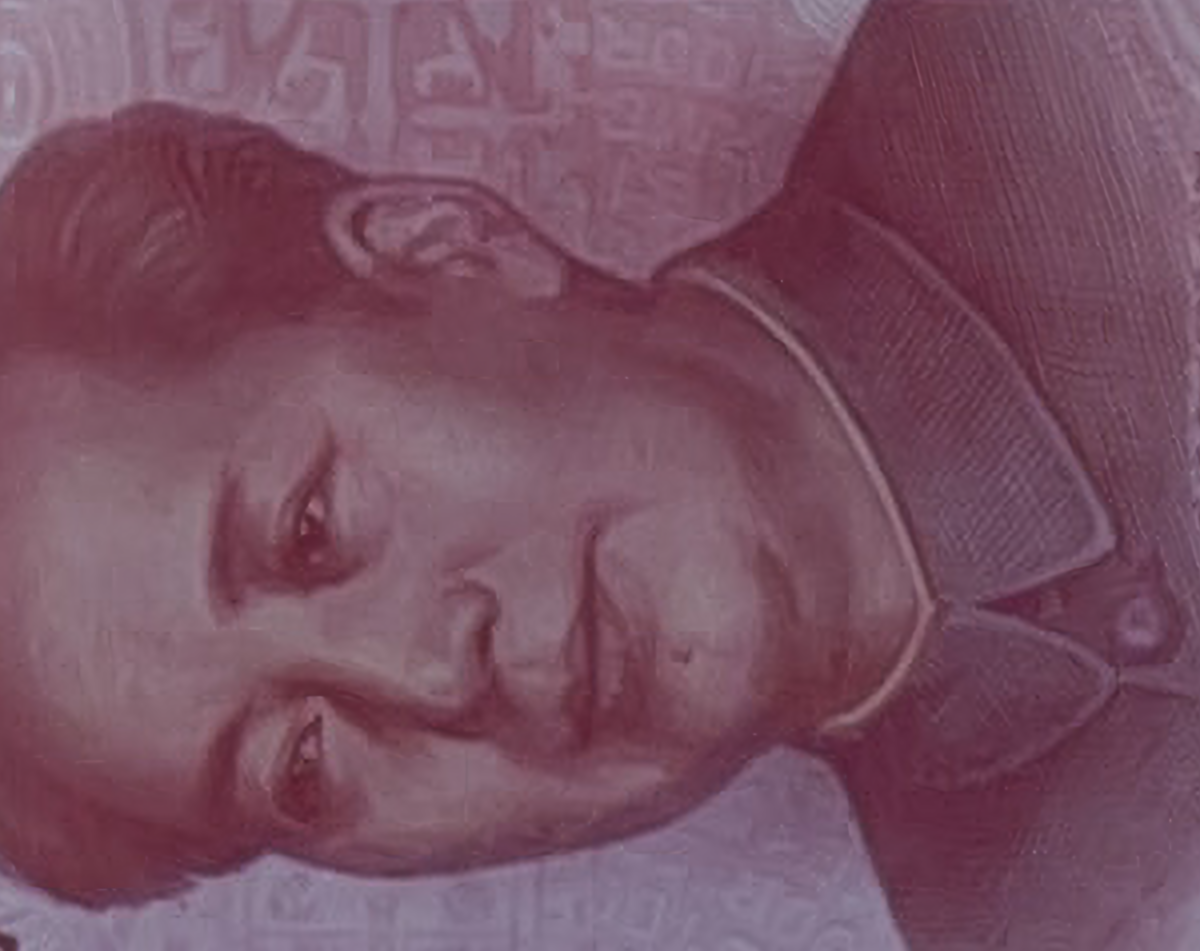} \hspace{-4mm} &
\includegraphics[width=0.18\textwidth]{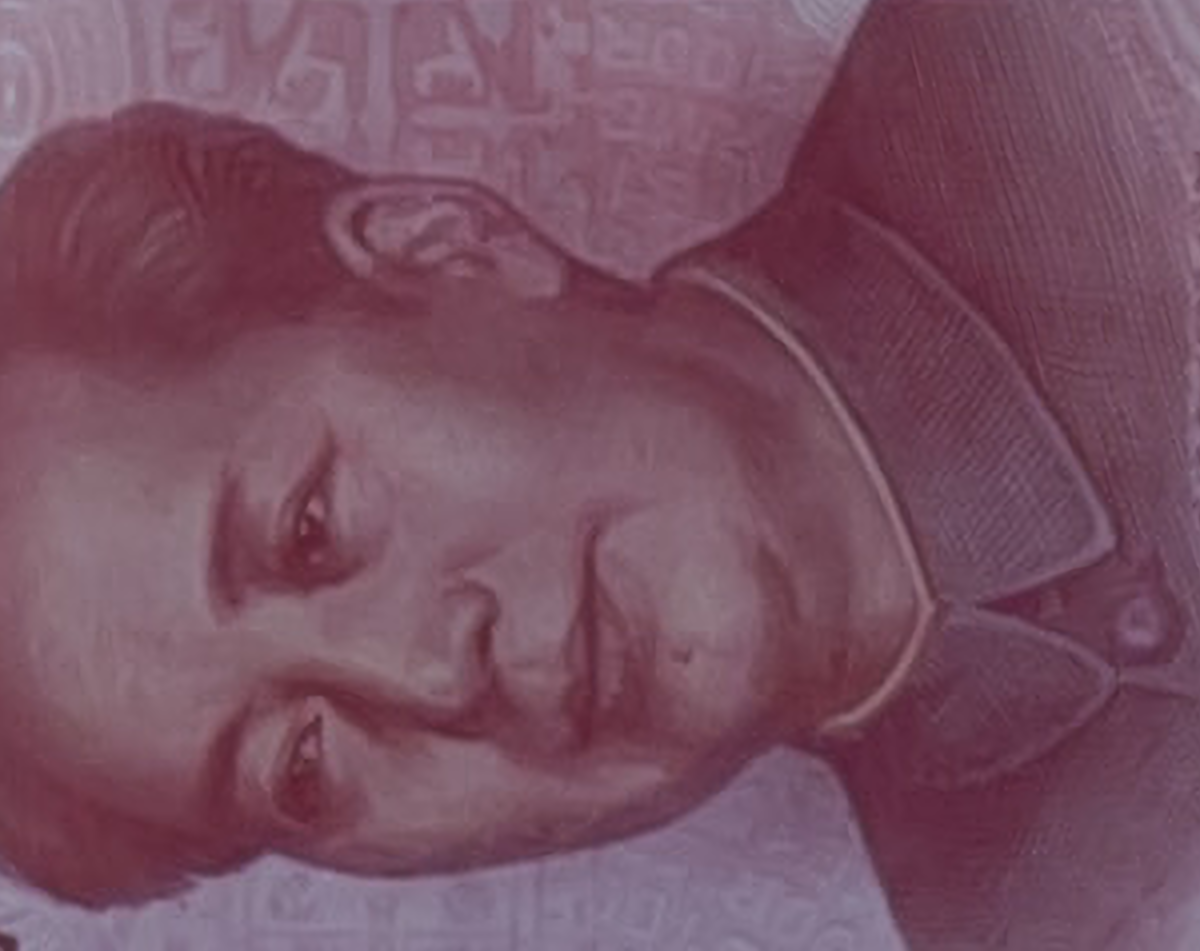} \hspace{-4mm} &
\includegraphics[width=0.18\textwidth]{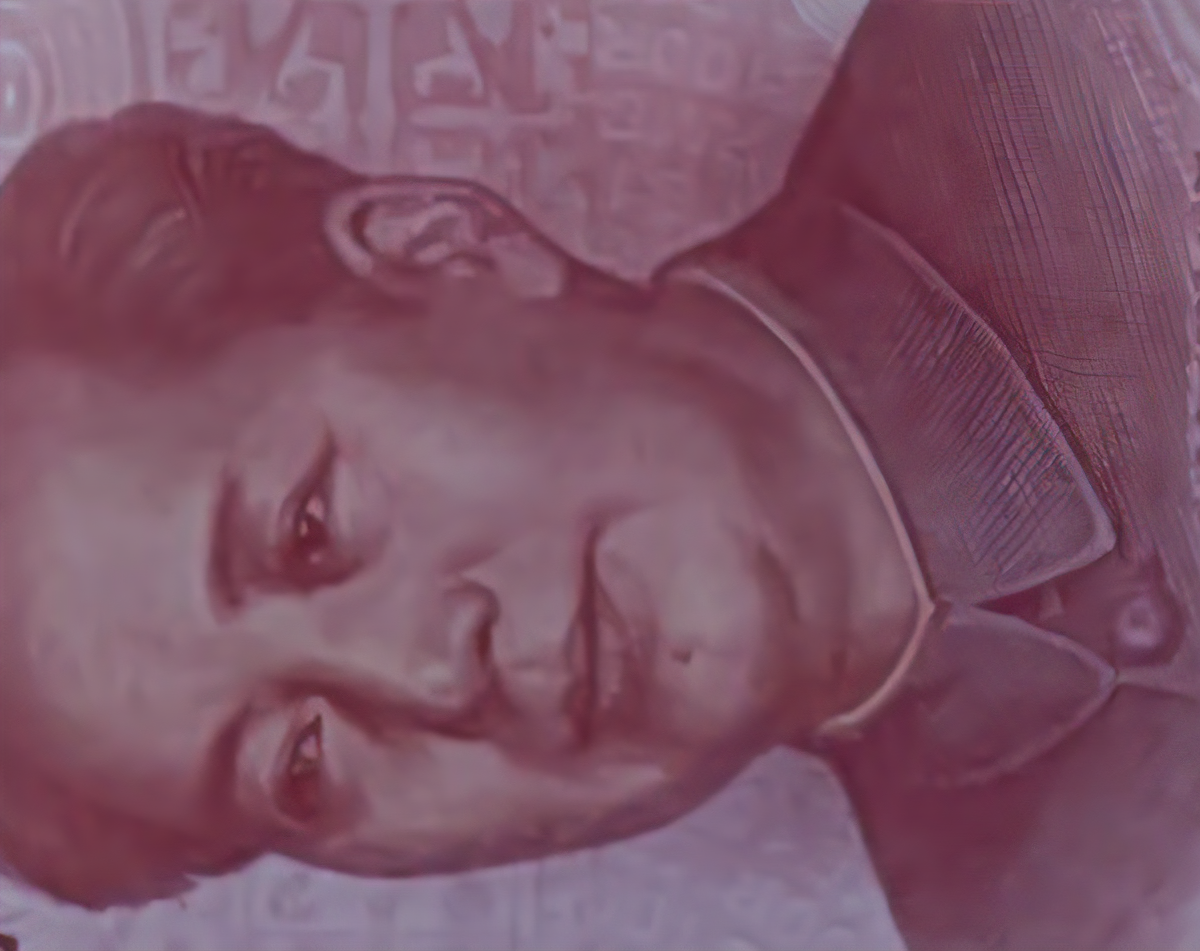} \hspace{-4mm} 
\\
Bicubic  \hspace{-4mm} &
IKC  \hspace{-4mm} &
DASR  \hspace{-4mm} &
FSSR  \hspace{-4mm}
\\
\includegraphics[width=0.18\textwidth]{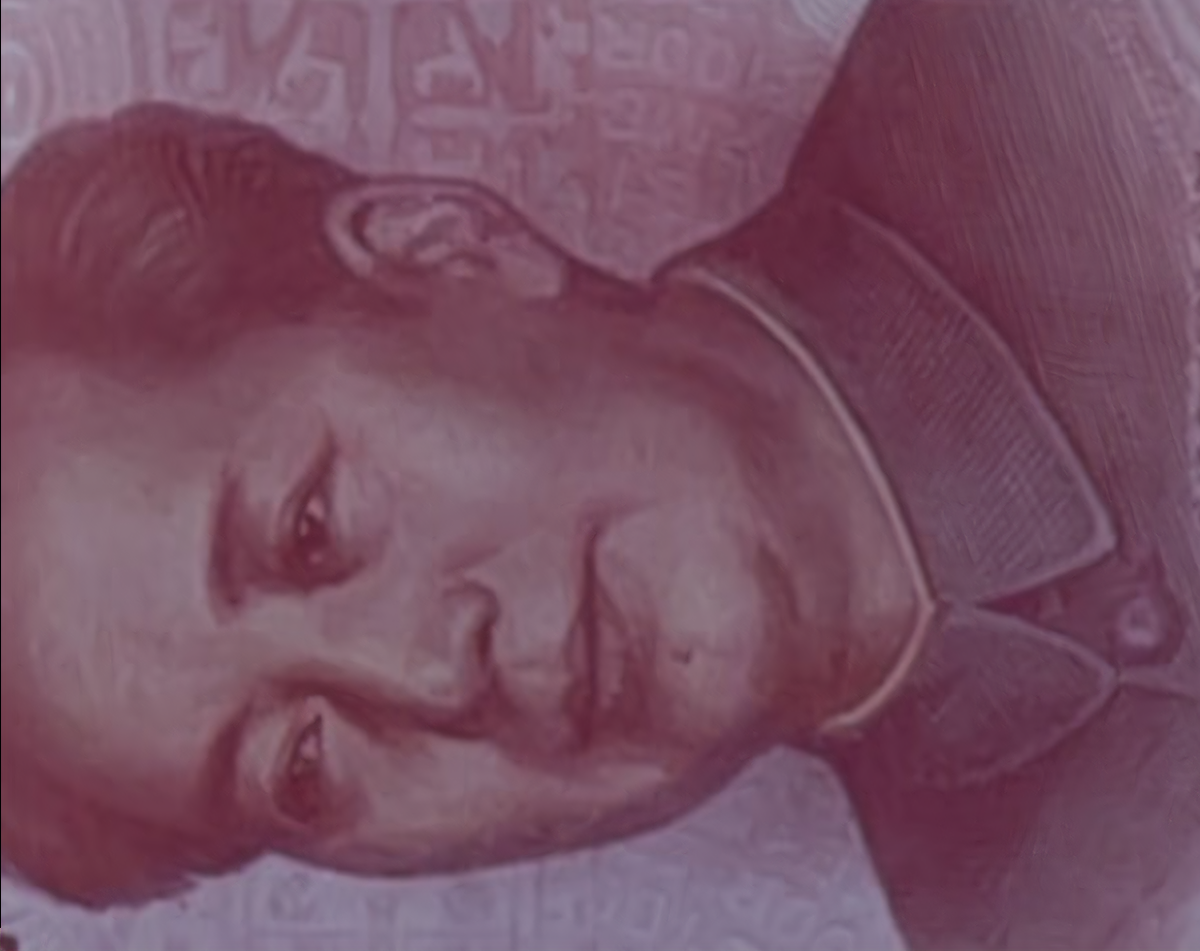} \hspace{-4mm} &
\includegraphics[width=0.18\textwidth]{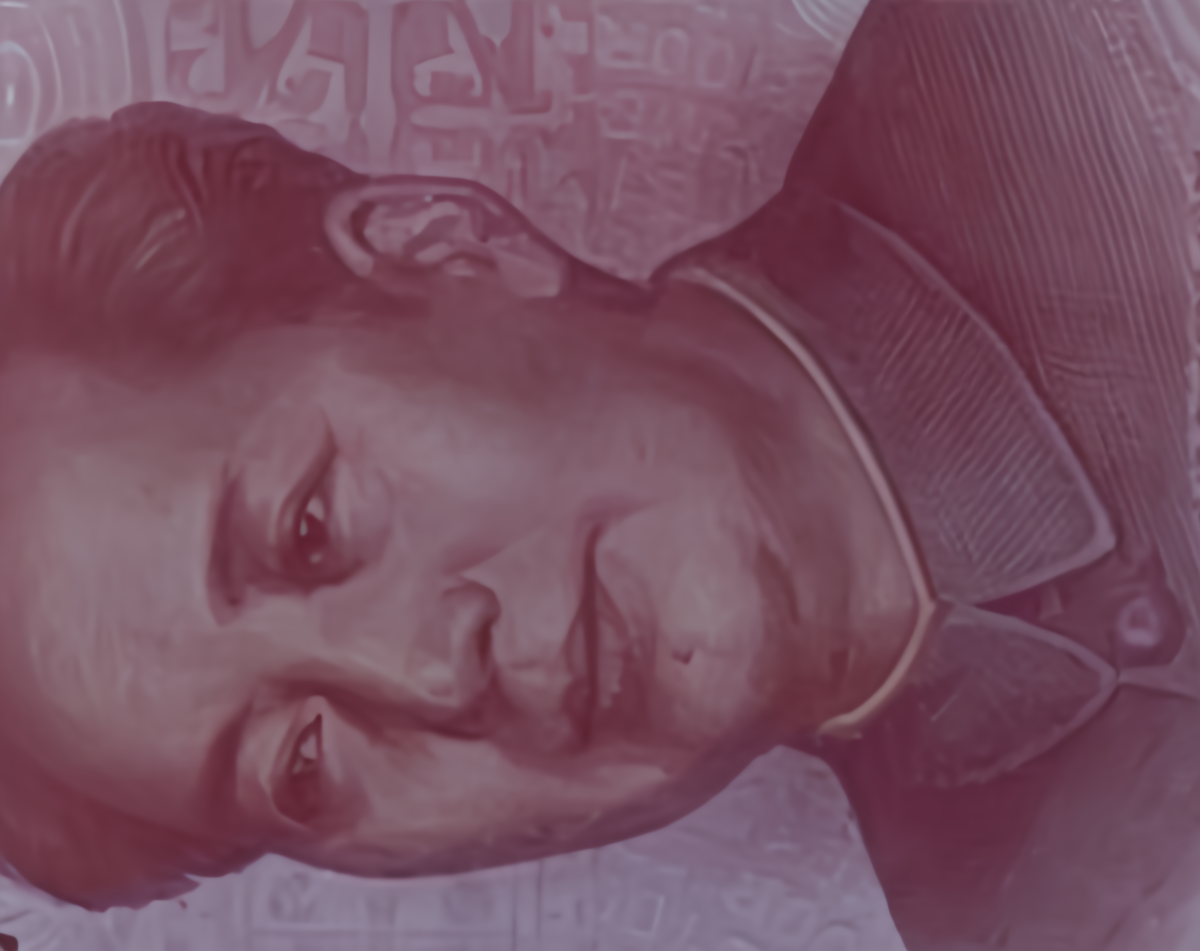} \hspace{-4mm} &
\includegraphics[width=0.18\textwidth]{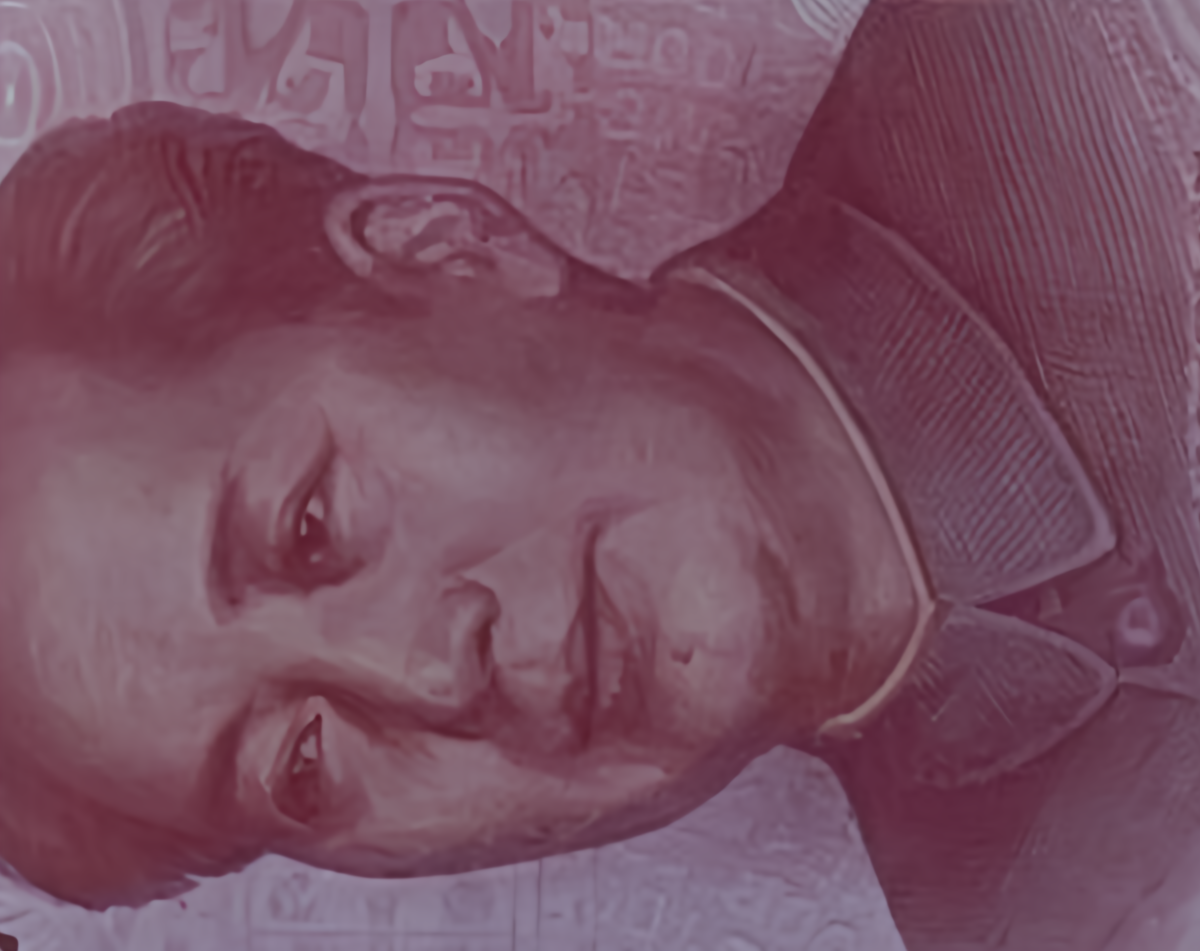} \hspace{-4mm} &
\includegraphics[width=0.18\textwidth]{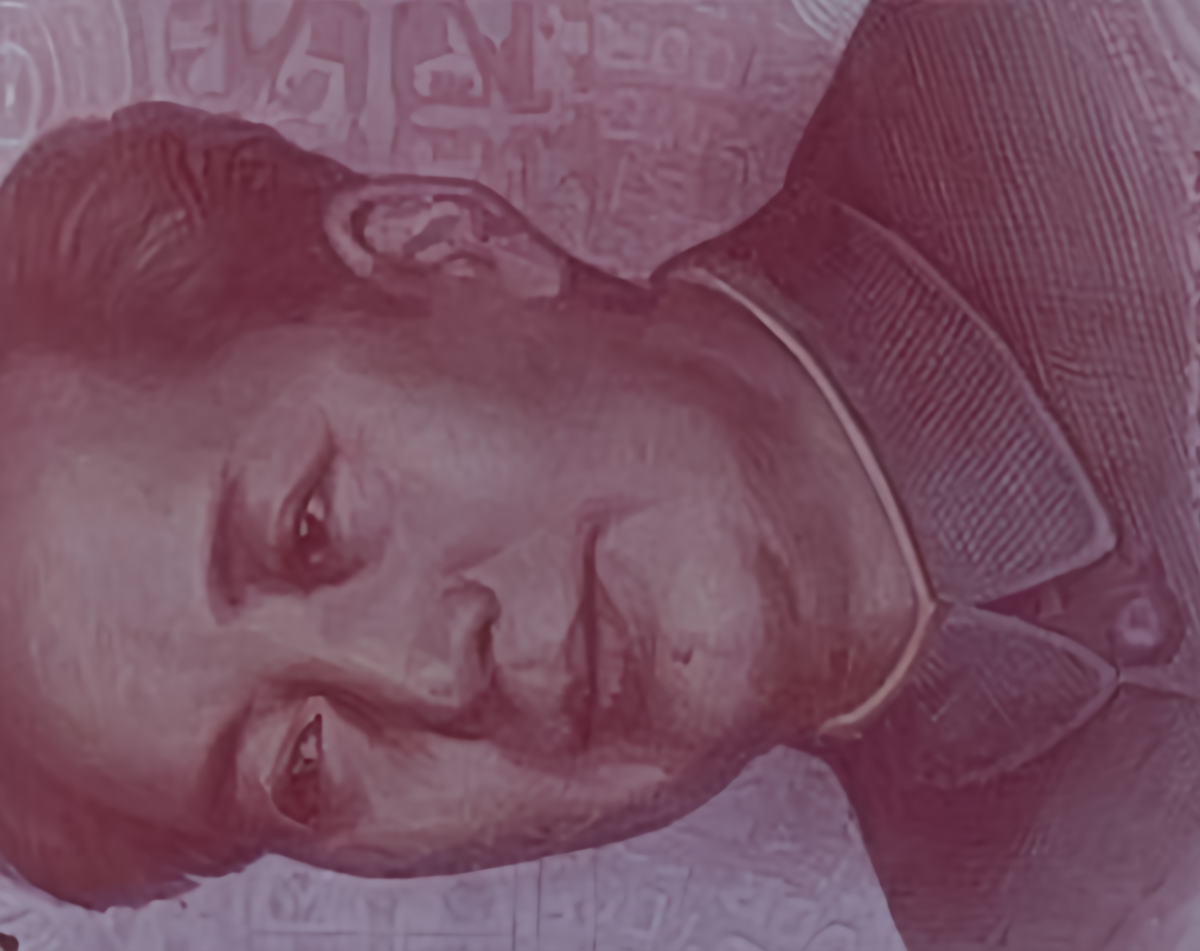} \hspace{-4mm}  
\\ 
USR-DA   \hspace{-4mm} &
BSRNet  \hspace{-4mm} &
Real-ESRNet \hspace{-4mm} &
Ours \hspace{-4mm}
\\
\end{tabular}
\end{adjustbox}
\\

% % % % one row

\end{tabular}}
\vspace{-3mm}
\caption{SR results of real-world images with scale factor $\times$4.}
\label{fig:realsupp2}
\vspace{-3mm}
\end{figure*}

%%%%%%%%%%%%%%%%%%%%%%%%%%%%%%%%%%%%%%%%%%%%%%%%%%%%%%%%%%%%%%%%%%%%%%%%%%%%%%%
%%%%%%%%%%%%%%%%%%%%%%%%%%%%%%%%%%%%%%%%%%%%%%%%%%%%%%%%%%%%%%%%%%%%%%%%%%%%%%%

\end{document}